\documentclass[letterpaper,twocolumn,10pt]{article}
\usepackage{style/usenix2019_v3}

\usepackage{color}
\usepackage{multirow}
\usepackage{amssymb}
\usepackage{url}

\usepackage{algorithmicx,algorithm}
\usepackage[noend]{algpseudocode}
\usepackage{subfigure}
\usepackage{enumitem}
\usepackage{times}
\usepackage{xspace}
\usepackage{latexsym}
\usepackage{amsmath}
\usepackage{booktabs}
\usepackage{tabularx}
\usepackage{balance}
\usepackage{graphicx}
\usepackage{svg}
\usepackage{pifont}
\svgsetup{
    inkscapepath=i/svg-inkscape/
}
\svgpath{{svg/}}
\usepackage{caption}
\usepackage{etoolbox}
\usepackage[all]{nowidow}

\setlist{nolistsep}
\newcommand{\parabf}[1]{\medskip\noindent\textbf{#1}}

\newcommand{\paraf}[1]{\noindent\textbf{#1}}
\newcommand{\cut}[1]{}

\newcommand{\sysname}{FastServe\xspace}

\newcommand{\revision}[1]{\textcolor{black}{#1}}

\newcommand{\revisionatc}[1]{\textcolor{black}{#1}}
\newcommand{\revisioneurosys}[1]{\textcolor{black}{#1}}

\newcommand{\revisionnsdi}[1]{\textcolor{black}{#1}}

\definecolor{ao}{rgb}{0.0, 0.5, 0.0}
\definecolor{method}{rgb}{0., 0., 0.}
\definecolor{cite}{rgb}{0.439, 0.141, 0.424}

\begin{document}
\title{Fast Distributed Inference Serving for Large Language Models}
\pagestyle{plain}
\author{
\rm{Bingyang Wu$^{*}$ \enskip
    Yinmin Zhong$^{*}$ \enskip
    Zili Zhang$^{*}$ \enskip
    Shengyu Liu \enskip
}
\vspace{1mm}
\\
\rm{
    Fangyue Liu \enskip
    Yuanhang Sun \enskip
    Gang Huang \enskip
    Xuanzhe Liu \enskip
    Xin Jin \enskip}
\vspace{3mm}
\\
{\textit{Peking University}}
}

\captionsetup{font=small}

\maketitle
{\let\thefootnote\relax\footnote{{$^*$Equal contribution.}}}

\begin{abstract}
Large language models (LLMs) power a new generation of interactive AI
applications exemplified by \mbox{ChatGPT}. The interactive nature of these
applications demands low latency for LLM inference. Existing
LLM serving systems use run-to-completion processing for inference jobs, which suffers
from head-of-line blocking and long latency.

We present \sysname, a distributed inference serving system for LLMs. \sysname
exploits the autoregressive pattern of LLM inference to enable preemption at the
granularity of each output token. \sysname uses preemptive scheduling to
minimize latency with a novel skip-join Multi-Level Feedback Queue scheduler.
Based on the new \emph{semi} information-agnostic setting of LLM inference, the
scheduler leverages the input length information to assign an appropriate
initial queue for each arrival job to join. The higher priority queues than the
joined queue are skipped to reduce demotions. We design an efficient GPU memory
management mechanism that proactively offloads and uploads intermediate state
between GPU memory and host memory for LLM inference. We build a system prototype of
\sysname and experimental results show
that compared to the state-of-the-art solution vLLM, \sysname improves
the throughput by up to 31.4$\times$ and
17.9$\times$ under the same average and tail latency requirements, respectively.
\end{abstract}

\section{Introduction}
\label{sec:introduction}

Advancements in large language models (LLMs) open new possibilities in a wide
variety of areas and trigger a new generation of interactive AI applications.
The most notable one is ChatGPT~\cite{chatgpt}
that enables users to interact with an AI agent in a
conversational way to solve tasks ranging from language translation to software
engineering. The impressive capability of ChatGPT makes it one of the fastest
growing applications in history~\cite{chatgptnews}. Many organizations follow the trend to
release LLMs and ChatGPT-like applications, such as the New Bing from
Microsoft~\cite{newbing}, Gemini from Google~\cite{gemini}, Claude-3~\cite{claude} from Anthropic, Qwen~\cite{qwen} from Alibaba, etc.

Inference serving is critical to interactive AI applications based on LLMs. In order to provide engaging user experience, the
interactive nature of these applications demands low latency for LLM inference. For example, users expect their inputs to ChatGPT to be responded instantly. Yet, the size and complexity
of LLMs put tremendous pressure on the underlying inference serving infrastructure.
Enterprises provision huge and expensive clusters that consist of accelerators like GPUs and TPUs to process LLM inference jobs.

\begin{figure}[t]
    \centering
    \includegraphics[width=\linewidth]{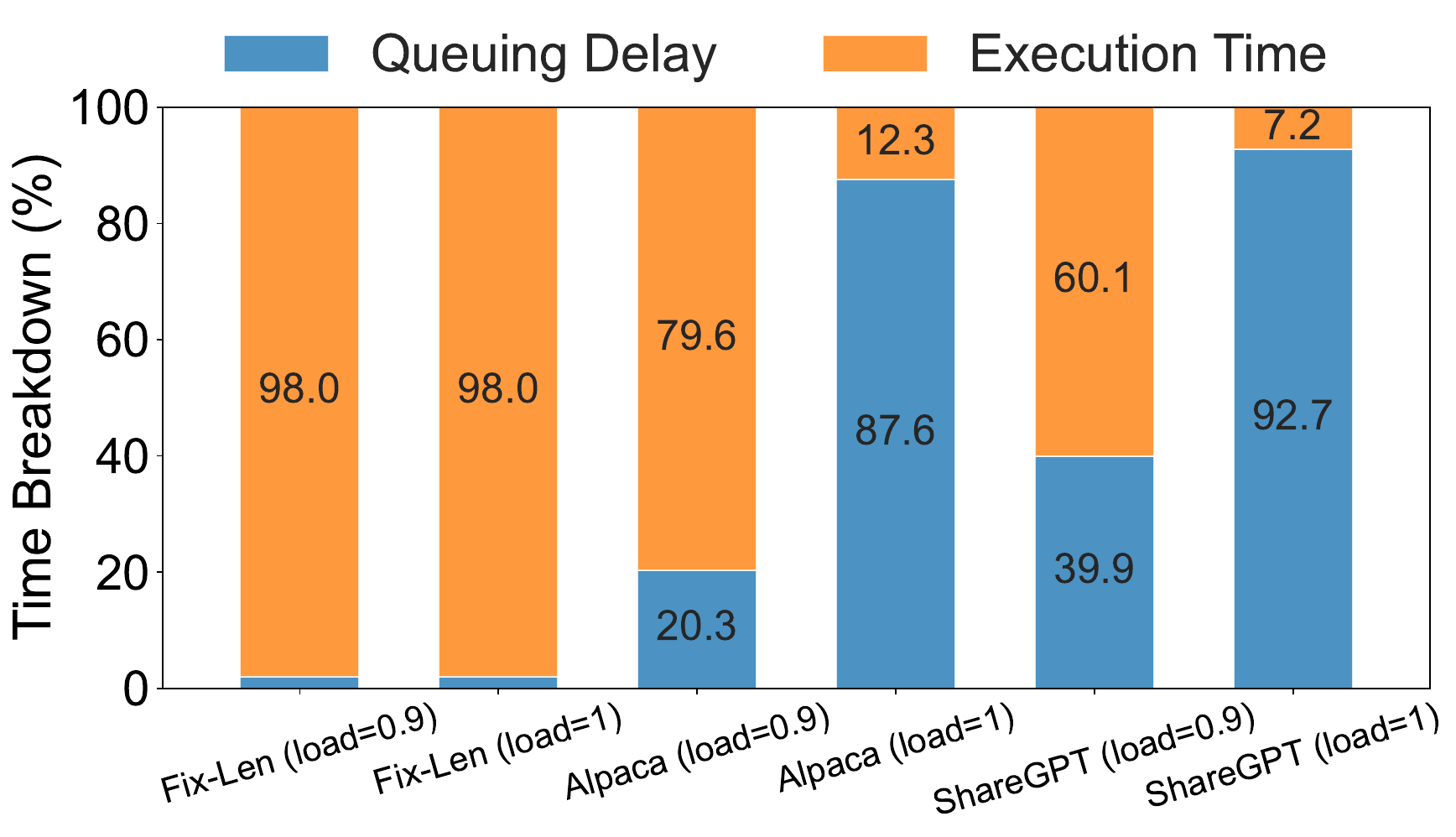}
    \vspace{-0.25in}
    \caption{Head-of-line blocking in LLM inference.}
    \vspace{-0.15in}
    \label{fig:intro:moti}
\end{figure}

LLM inference has its own unique characteristics (\S\ref{sec:background}) that
are different from other deep neural network (DNN) model inference like
ResNet~\cite{he2016deep}. DNN inference jobs are typically deterministic and
highly-predictable~\cite{gujarati2020serving}, i.e., the execution time of an
inference job is mainly decided by the model and the hardware. For example,
different input images have similar execution time on the same ResNet model on a
given GPU. In contrast, LLM inference jobs have a special \emph{autoregressive}
pattern. An LLM inference job contains multiple iterations. Each iteration
generates one output token, and each output token is appended to the input to
generate the next output token in the next iteration. The execution time depends
on both the input length and the output length, the latter of which is not known
\emph{a priori}.

Existing inference serving solutions like Clockwork~\cite{gujarati2020serving} and
Shepherd~\cite{zhangshepherd} target deterministic model
inference like ResNet~\cite{he2016deep}. They rely on accurate execution
time profiling to schedule jobs, which do not work for LLM inference
that has variable execution time.
Orca~\cite{yu2022orca}, designed for LLM inference, introduces iteration-level scheduling
that dynamically adds new jobs or removes completed ones at the end of each iteration.
vLLM~\cite{vllm} further introduces
PagedAttention to reduce the memory fragmentation of the intermediate state of LLM inference jobs.
However, they both use first-come-first-served (FCFS) to process 
inference jobs. Once a job is scheduled,
it runs until it finishes. Due to the limited GPU memory and the strict latency requirement,
the current processing batch cannot be expanded with an arbitrary number of incoming jobs,
thus a long job may block the incoming ones, known as head-of-line blocking~\cite{shinjuku}. The problem is
particularly acute for LLM inference jobs. A large LLM inference job, i.e., with long input and
output length, would run for a long time to block incoming short jobs.   

Figure~\ref{fig:intro:moti} demonstrates the problem based on real-world datasets. The
detailed setup is in~\S\ref{sec:evaluation}. Ideally, if the input length and output length of
the jobs are all the same among jobs, \revision{there is almost no queuing delay even when the load reaches capacity ($load\approx1$)},
as the first two columns in Figure~\ref{fig:intro:moti} suggest. However,
LLM datasets like ShareGPT~\cite{sharegpt} and Alpaca~\cite{alpaca} show that 
real-world workloads are highly skewed. The long-tail distribution of output
length leads to a long queuing delay. Figure~\ref{fig:intro:moti} shows that
up to 90\% of total latency is the queuing delay for real-world datasets. In this case, optimizing
execution time is not enough, because it only contributes to a small portion of
the end-to-end latency. Instead, we need to optimize the queuing delay, which is the
major contributor to the end-to-end latency.

We present \sysname, a distributed inference serving system for LLMs. \sysname exploits the
autoregressive pattern of LLM inference and iteration-level scheduling to enable preemption at the granularity
of each output token. Specifically, when one scheduled job finishes
generating an output token, \sysname can decide whether to continue this job or
preempt it with another job in the queue. This allows \sysname to use preemptive
scheduling to eliminate head-of-line blocking problem and minimize latency.

The core of \sysname is a novel \emph{skip-join} Multi-Level Feedback Queue (MLFQ)
scheduler. MLFQ is a classic approach to minimize latency in
information-agnostic settings~\cite{bai2015information}. Each job first enters
the highest priority queue, and is demoted to the next priority queue if it does
not finish after a threshold. The key difference between LLM inference and the
classic setting is that LLM inference is \emph{semi} information-agnostic, i.e.,
while the output length is not known \emph{a priori}, the input length is known.
Because of the autoregressive pattern of LLM inference, the input length decides
the execution time to generate the first output token, which can be
significantly larger than those of the later tokens (\S\ref{subsec:mlfq}). For a
long input and a short output, the execution time of the first output token
dominates the entire job. We leverage this characteristic to extend the classic
MLFQ with skip-join. Instead of always entering the highest priority queue, each
arrival job joins an appropriate queue by comparing its execution time of the
first output token with the quantum of the queues. The higher priority queues
are skipped to reduce demotions.

Preemptive scheduling introduces extra memory overhead to maintain
intermediate state for started but unfinished jobs. LLMs maintain a key-value
cache for each Transformer layer to store intermediate state (\S\ref{subsec:background-serving-systems}). In FCFS, the
cache only needs to store the intermediate state of the scheduled jobs in the processing batch, limited by the maximum batch size. But in MLFQ, more jobs may have started but are demoted to lower
priority queues. The cache has to maintain the intermediate state for all started but unfinished jobs in MLFQ. The cache can overflow, given the large
size of intermediate state and the limited memory capacity of GPUs. Naively, the scheduler can
pause starting new jobs when the cache is full, but this again introduces
head-of-line blocking. Instead, we design \emph{proactive} GPU memory management mechanism that
proactively offloads the state of the jobs in low-priority queues to the host
memory when the cache is close to full, and uploads the state back when these jobs
are to be scheduled. We use pipelining and asynchronous memory operations to
improve the efficiency.

For large models that do not fit in one GPU, \sysname leverages parallelization
strategies including tensor parallelism~\cite{shoeybi2020megatronlm} and pipeline parallelism~\cite{huang2019gpipe} to perform distributed
inference serving with multiple GPUs
(\S\ref{subsec:design:distributed_support}). The scheduler runs multiple batches of jobs
concurrently in a pipeline to minimize pipeline bubbles.
The key-value cache manager partitions the key-value cache over multiple GPUs,
and handles swapping between GPU memory
and host memory in a distributed manner.

We implement a system prototype of \sysname and integrate many optimization techniques like PagedAttention~\cite{vllm}. We evaluate \sysname on different configurations of
LLMs with real-world LLM inference workloads. In particular, we evaluate the end-to-end performance of
\sysname for OPT-175B~\cite{zhang2022opt} (an open-source LLM similar to the largest GPT-3 model) on 16 NVIDIA A100
GPUs. The
experiments show that compared to the state-of-the-art solution vLLM~\cite{vllm}, \sysname
improves the throughput by up to 31.4$\times$ and
17.9$\times$ under the same average and tail latency requirements, respectively.

\section{Background and Motivation}
\label{sec:background}

\begin{figure}[t]
    \centering
    \includegraphics[width=0.88\columnwidth]{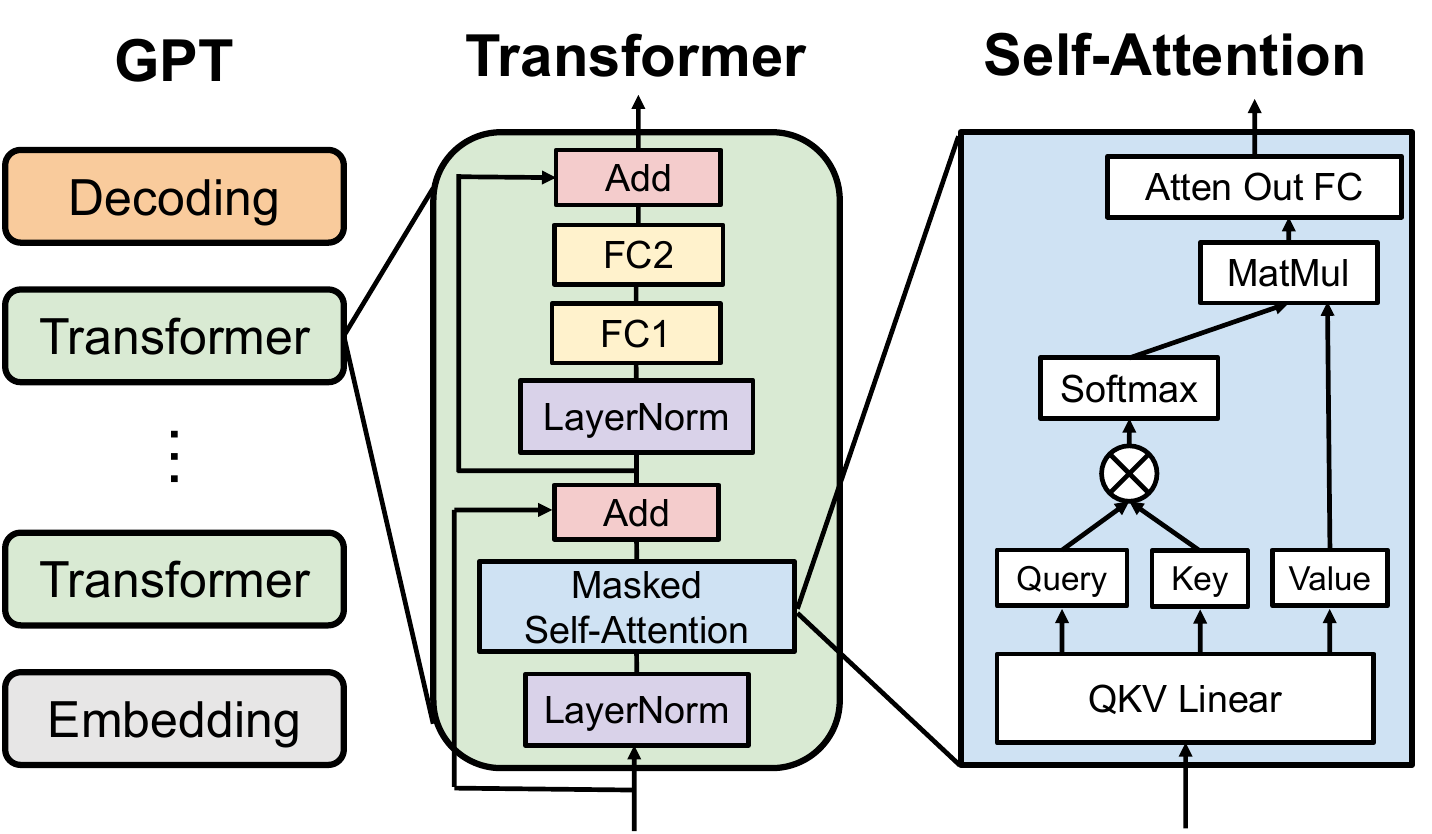}
    \vspace{-0.05in}
    \caption{GPT-like model architecture.}
    \vspace{-0.15in}
    \label{fig:transformer}
\end{figure}

\subsection{LLM Inference and Applications}
\paraf{LLM inference.} The LLM family~\cite{brown2020language,zhang2022opt,touvron2023llama} comprises a set of
language models built on the foundation of Transformer~\cite{vaswani2017attention}.
LLM inference operates in an autoregressive fashion, where the input,
often known as a prompt, is processed as a sequence of tokens. It then produces
a probability distribution for the succeeding token to be selected. This processing
and selection mechanism for each output token is referred to as an \emph{iteration}.
Once trained with a vast corpus, LLM is capable of executing high-quality language tasks.

For instance, given the input "knowledge is", it can assign a higher probability
to "power" than to "apple". The first output token is appended
to the initial prompt and fed into LLM for subsequent token generation. This process
continues until a unique <EOS> token, symbolizing the end of sequence, is generated,
or a predetermined maximum output length is reached. This inference process markedly
differs from those of other models like ResNet, where execution time is usually
deterministic and predictable~\cite{gujarati2020serving}. While each iteration's
execution maintains these characteristics in the LLM model, the number of iterations
(i.e., output length) is variable, resulting in an unpredictable total inference job execution time.

\parabf{LLM applications.} 
The LLMs' primary commission is to predict the next token for an input prompt.
Utilizing prompt engineering~\cite{brown2020language}, downstream NLP tasks can be reformulated as generation
tasks based on LLMs.
Specifically, for a translation task, one can start the prompt by adding
"Translate the following English text into French text" before the original text.
By doing so, LLM will be guided to generate the desired translated French text in response.

ChatGPT~\cite{chatgpt} is a representative LLM-based application.
After supervised fine-tuning for the
conversational task and an alignment procedure using Reinforcement Learning from
Human Feedback (RLHF) on the original GPT model~\cite{brown2020language}, ChatGPT facilitates
interactive conversations with an AI agent, allowing users to address a wide range of tasks.
These tasks include translation, question-answering, summarization, as well as more intricate undertakings
like sentiment analysis, creative writing, and domain-specific problem-solving.
Despite its remarkable capabilities, the interactive nature of ChatGPT imposes a
considerable pressure on the underlying inference serving infrastructure.
Due to the need for rapid responses,
keeping low latency is crucial for ensuring the performance
of ChatGPT-like interactive applications.

\subsection{Inference Serving Systems}
\label{subsec:background-serving-systems}
Most existing inference serving systems, such as Tensorflow Serving~\cite{olston2017tensorflow}
and Triton Inference Server~\cite{nvidiatriton}, are agnostic to DNN models. They serve as an
abstraction above the underlying execution engine, which queues the arriving jobs, dispatches jobs to available
computing resources, and returns the results to clients. To fully utilize the GPUs,
they typically batch jobs together for parallel processing. With batching, the input tensors
from multiple jobs are concatenated and fed into the model as a whole. \revision{Despite better utilization, the drawback of batching is higher memory overhead.
The substantial size of LLMs and the huge intermediate state restricts the maximum batch
size for LLM inference~\cite{vllm}.}

\begin{figure}[t]
    \centering
    \includegraphics[width=0.84\linewidth]{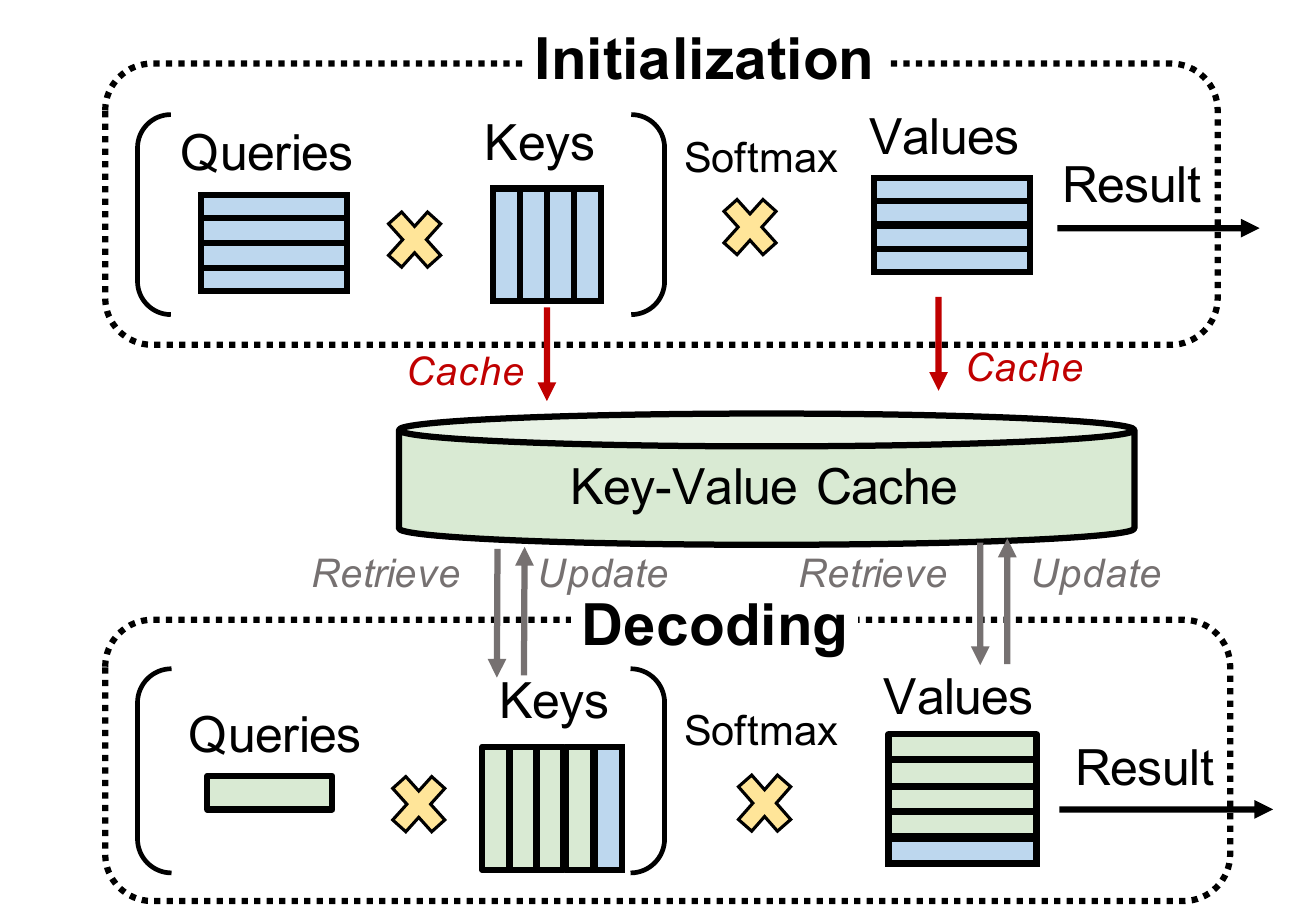}
    \vspace{-0.03in}
    \caption{Demonstration of KV cache.}
    \vspace{-0.15in}
    \label{fig:kvcache}
\end{figure}

As the popularity of LLMs rapidly increases, inference serving systems have evolved to include optimizations specific to the unique
architecture and iterative generation pattern of LLMs. The major part of LLM's architecture is a stack of Transformer layers,
as shown in Figure~\ref{fig:transformer}. 
In a Transformer layer, the Masked Self-Attention module is the core component that
distinguishes it from other architectures like CNNs. During each iteration of LLM inference, for each token, the attention operator
requires the \textit{keys} and \textit{values} of preceding tokens. A naive, stateless
implementation always recomputes all the preceding keys and values in each iteration. To
avoid such recomputation overhead, fairseq~\cite{ott2019fairseq} suggests saving
the keys and values in a \emph{key-value cache} across iterations. This optimization divides the inference procedure into two distinct phases: the \textit{initialization phase}
and the \textit{decoding phase}. Figure~\ref{fig:kvcache} demonstrates the
key-value cache usage in both phases. During the \textit{initialization phase}, which corresponds
to the first iteration, the LLM generates the key-value cache for each token in the input prompt.
In the subsequent \textit{decoding phase}, the LLM only needs to compute the query, key, and value of
one newly generated token, leveraging the precomputed key-value cache to facilitate the process step by step.
Consequently, the execution time of iterations in the decoding phase is typically smaller compared
to that of the initialization phase, i.e., the first iteration. It is worth noting that other
Transformer-based systems, such as HuggingFace~\cite{wolf2020huggingfaces} and
FasterTransformer~\cite{fastertransformer}, also incorporate this optimization technique,
leading to improved efficiency during inference.

Another important optimization is iteration-level scheduling proposed by Orca~\cite{yu2022orca}.
Naive job-level scheduling executes a batch of jobs until all jobs finish. The early finished jobs cannot
be returned to the clients immediately, while newly arrived jobs have to wait until the ongoing batch completes processing.
However, with iteration-level scheduling, the execution engine executes only a single iteration on the batch
at a time, generating one output token for each job. After each iteration, completed jobs leave the batch,
and newly arrived jobs can join in. Nevertheless, the GPU memory capacity limits the maximum batch size,
and the strict service-level-objects (SLOs) of interactive applications also play a role in determining the
appropriate batch size.
vLLM~\cite{vllm} further improves the efficiency of LLM inference by introducing PagedAttention, which allocates the key-value
cache gradually in block-grained during inference instead of allocating for the maximum output length at the beginning.

\subsection{Opportunities and Challenges}
\label{subsec:challenges}

\paraf{Opportunity: preemptive scheduling.} The major limitation of existing inference serving
systems for LLMs~\cite{fastertransformer,yu2022orca} is their reliance on simple FCFS (First-Come-First-Serve)
scheduling and run-to-completion execution. As shown in Figure~\ref{fig:intro:moti}, this approach leads to severe head-of-line blocking. 
Queuing delay contributes up to 90\% of the total latency in real workload, which
significantly impacts the performance of LLM inference. To overcome this challenge, preemptive scheduling
can be employed. In LLM inference, each job comprises multiple iterations, with each iteration generating
one output token. The opportunity lies in exploiting this autoregressive pattern to enable
preemption at the iteration level, meaning that one job can be preempted when it finishes
generating an output token for another job. Leveraging preemption capability,
the scheduler can employ preemptive scheduling policies to prevent head-of-line blocking and optimize average latency.
Nevertheless, preemptive scheduling presents two challenges for the existing LLM inference system.

\parabf{Challenge 1: variable job size.} 
Shortest Remaining Processing Time (SRPT)~\cite{schrage1968proof} is a widely-used preemptive scheduling policy to minimize average latency.
However, applying SRPT to LLM inference presents challenges
due to the iterative nature of LLMs. Unlike one-shot prediction tasks like image classification,
LLM inference involves multiple iterations. While the execution time for one iteration
(generating one output token) can be determined based on the model architecture and hardware,
the total number of iterations (i.e., the output sequence length) remains unknown and is
challenging to predict since it depends on the semantics of the job.
Real-world datasets collected from conversations with LLMs, like ShareGPT~\cite{sharegpt} and Alpaca~\cite{alpaca}, exhibit a long-tailed distribution of the output length and input length~\cite{vllm}.
Consequently, SRPT cannot be directly employed for LLM inference to minimize the average latency.

\parabf{Challenge 2: GPU memory overhead.}
Preemptive scheduling policies introduce additional GPU memory consumption during LLM inference,
unlike FCFS with run-to-completion, which only needs to maintain the key-value cache for ongoing jobs.
In contrast, preemptive scheduling must keep the key-value cache in GPU memory for all preempted jobs
in the pending state, to be used for future token generation. This key-value cache consumes a substantial
amount of GPU memory, leading to potential challenges. For instance, a single job of OPT 175B with
an input sequence length of 512 requires at least 2.3 GB of memory for the key-value cache (\S\ref{subsec:kvcache}).
Due to scarce GPU memory capacity, the size of the key-value cache becomes a critical factor
affecting the effectiveness of preemptive scheduling policies.
Prior works have proposed memory-saving techniques for the
key-value cache. Multi-Query Attention~\cite{shazeer2019fast} and Group-Query Attention~\cite{gqa} try to
reduce the memory consumption by sharing key-value tensors between attention heads. They may hurt the capability of LLMs and 
the memory consumption still grows linearly with the sequence length. vLLM~\cite{vllm} manages the key-value cache at the granularity
of blocks to reduce GPU memory fragmentation. It cannot reduce the
memory usage caused by the key-value cache itself. As the context length
increases~\cite{longchat2023}, the memory consumption of the key-value cache is a hard problem and is becoming increasingly important.
\section{\sysname Overview}

\begin{figure}[t]
    \centering
    \includegraphics[width=1.0\columnwidth]{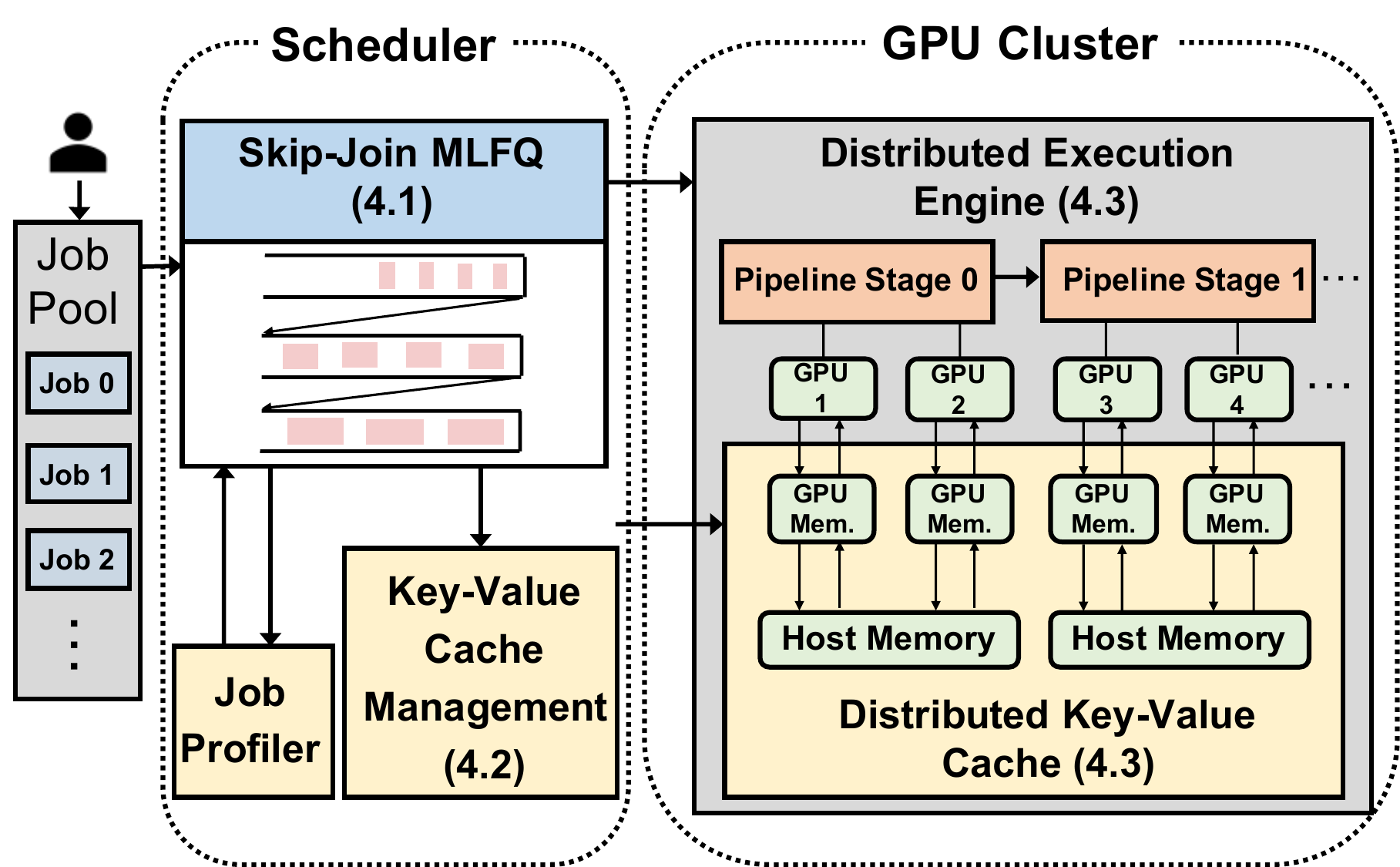}
    \vspace{-0.15in}
    \caption{\sysname architecture.}
    \vspace{-0.15in}
    \label{fig:architecture}
\end{figure}

\subsection{Desired Properties}

LLMs come with unique characteristics that pose challenges to distributed
computation and GPU memory consumption. Our goal is to develop an efficient
inference serving system for LLMs that fulfills the following three requirements.

\begin{itemize}[leftmargin=*]
    \item \textbf{Low latency and high throughput.} Our focus centers on interactive LLM applications, where users have high expectations for fast response. To measure it quantitatively, we want the maximum throughput as high as possible under a certain latency requirement.

    \item \textbf{Efficient GPU memory management.} LLMs pose a significant challenge in terms of
    GPU memory consumption, which necessitates an effective GPU memory management approach for both the
    model and intermediate states.

    \item \textbf{Scalable distributed execution.} The nature of LLMs demands multiple GPUs to enable
    distributed inference effectively, which requires the system to support scalable distributed execution cross GPU servers.
\end{itemize}

\subsection{Overall Architecture}

Figure~\ref{fig:architecture} shows the architecture of \sysname.
Jobs are submitted to the job pool. The scheduler
utilizes information from the job profiler to determine the initial job priority and then places the job
in the skip-join MLFQ (\S\ref{subsec:mlfq}) to mitigate head-of-line blocking.

For execution, the scheduler picks the jobs based on their priority within the skip-join MLFQ to form a pre-defined maximum batch size and dispatches the batch to the distributed execution engine 
to perform one iteration. The distributed
execution engine collaborates with the distributed key-value cache
to access and update the key-value tensors relevant to the respective job.
To tackle the challenge of limited GPU memory capacity, the key-value cache manager proactively swaps key-value tensors between GPU memory and host memory (\S\ref{subsec:kvcache}).

To accommodate extreme large models such as OPT-175B, \sysname employs distributed inference, enabling both tensor parallelism and pipeline parallelism. \sysname incorporates extensions into the 
scheduler and key-value cache to enable seamless support for distributed execution (\S\ref{subsec:design:distributed_support}).

\section{\sysname Design}
\label{sec:design}

In this section, we first introduce the skip-join MLFQ scheduler
to minimize latency (\S\ref{subsec:mlfq}). Then, we present a proactive KV cache management
mechanism designed to effectively ameliorate the GPU memory capacity
constraint (\S\ref{subsec:kvcache}). Finally, we demonstrate how to apply these techniques to the
distributed settings (\S\ref{subsec:design:distributed_support}).

\begin{figure}[!t]
    \centering
    \includegraphics[width=0.8\linewidth]{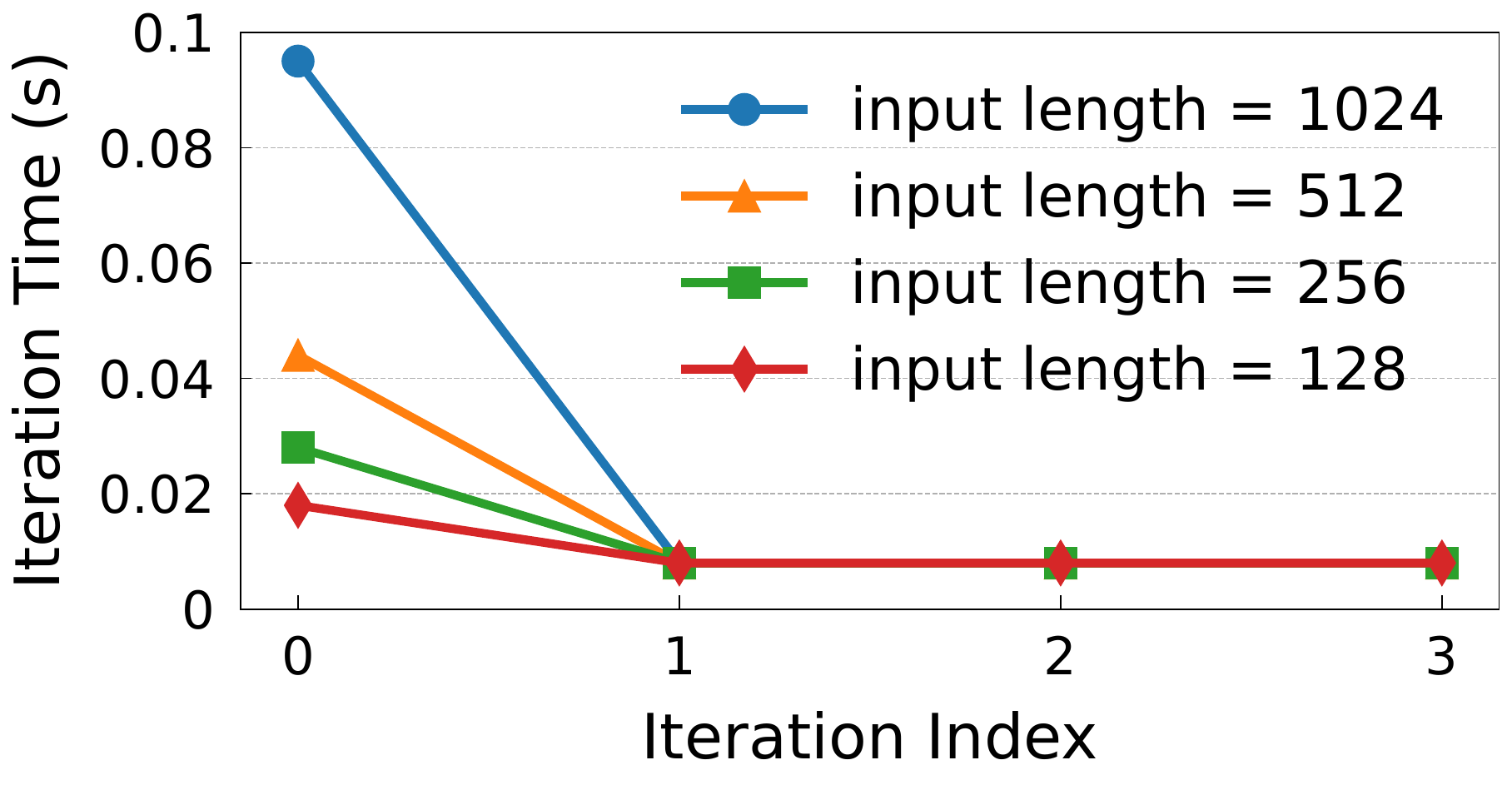}
    \vspace{-0.1in}
    \caption{Execution time of the first four iterations.}
    \vspace{-0.15in}
    \label{fig:design:iteration_time}
\end{figure}

\subsection{Skip-Join MLFQ Scheduler}
\label{subsec:mlfq}
\paraf{Strawman: fixed priority scheduling.}
\revision{To support preemptive scheduling, we need a priority-based 
scheduler to decide which jobs to preempt and which to execute.}
One naive solution is to assign a fixed priority to
each job based on its input length. In this case, when
the initialization phase dominates the total latency, the
fixed priority scheduling can approximate the optimal
performance as the SRPT policy does. However, although this
solution leverages the information about the initialization phase,
it ignores the characteristics of the decoding phase. Many real-world
datasets like ShareGPT and Alpaca show a long tail distribution
implying that jobs with a long output length also exist. 
When the decoding phase dominates the total latency, the fixed
priority scheduling may deviate from the optimal performance of SRPT.

\parabf{Strawman: naive MLFQ.}
Due to the indeterminate job size of LLM inference, directly applying SRPT
is not feasible. In information-agnostic settings, Least-Attained Service
(LAS) has been shown to approximate SRPT effectively.
Due to the job switching overhead of LAS, the practical approach is Multi-Level
Feedback Queue (MLFQ) which has gained popularity in various scheduling
systems~\cite{flowpreempt, alizadeh2013pfabric, chowdhury2015efficient, bai2015information, gu2019tiresias}.
MLFQ operates multiple queues, each with a different priority level. Upon arrival,
a job enters the highest priority queue and gets demoted to the next level queue if its execution time exceeds
a quantum. The value of quantum is a tunable parameter assigned to each queue,
e.g., higher priority queues typically have shorter quantum values.

Although MLFQ assumes no prior knowledge of the job size, it is not well suited
for LLM serving.
Figure~\ref{fig:design:iteration_time} shows the iteration time
of OPT 2.7B on an NVIDIA A100, varying the input sequence length. Notably,
the initialization phase (i.e., the first iteration) time
exceeds the decoding phase duration. As the input sequence
length increases, so does the initialization phase time. This behavior can be attributed
to the key-value cache optimization (\S\ref{subsec:background-serving-systems}).
During the first iteration, computations for all key-value tensors of input tokens
are performed and cached. In subsequent iterations, only one newly generated token's
key-value tensors are computed, and the rest are retrieved from the cache.

When employing the original MLFQ, a job is immediately assigned to the highest
priority queue upon arrival. However, due to its substantial initialization
phase time, the job may deplete its quantum before completing its first iteration.
This situation presents a scheduling dilemma.
If the scheduler preempts the job, intermediate activations are dropped and recomputed later,
resulting in a waste of valuable computing resources and time.
On the other hand, if the scheduler chooses not to preempt the job,
it violates the fundamental design purpose of MLFQ and potentially
suffers from head-of-line blocking once again.

\begin{figure}[!t]
    \centering
    \includegraphics[width=0.8\linewidth]{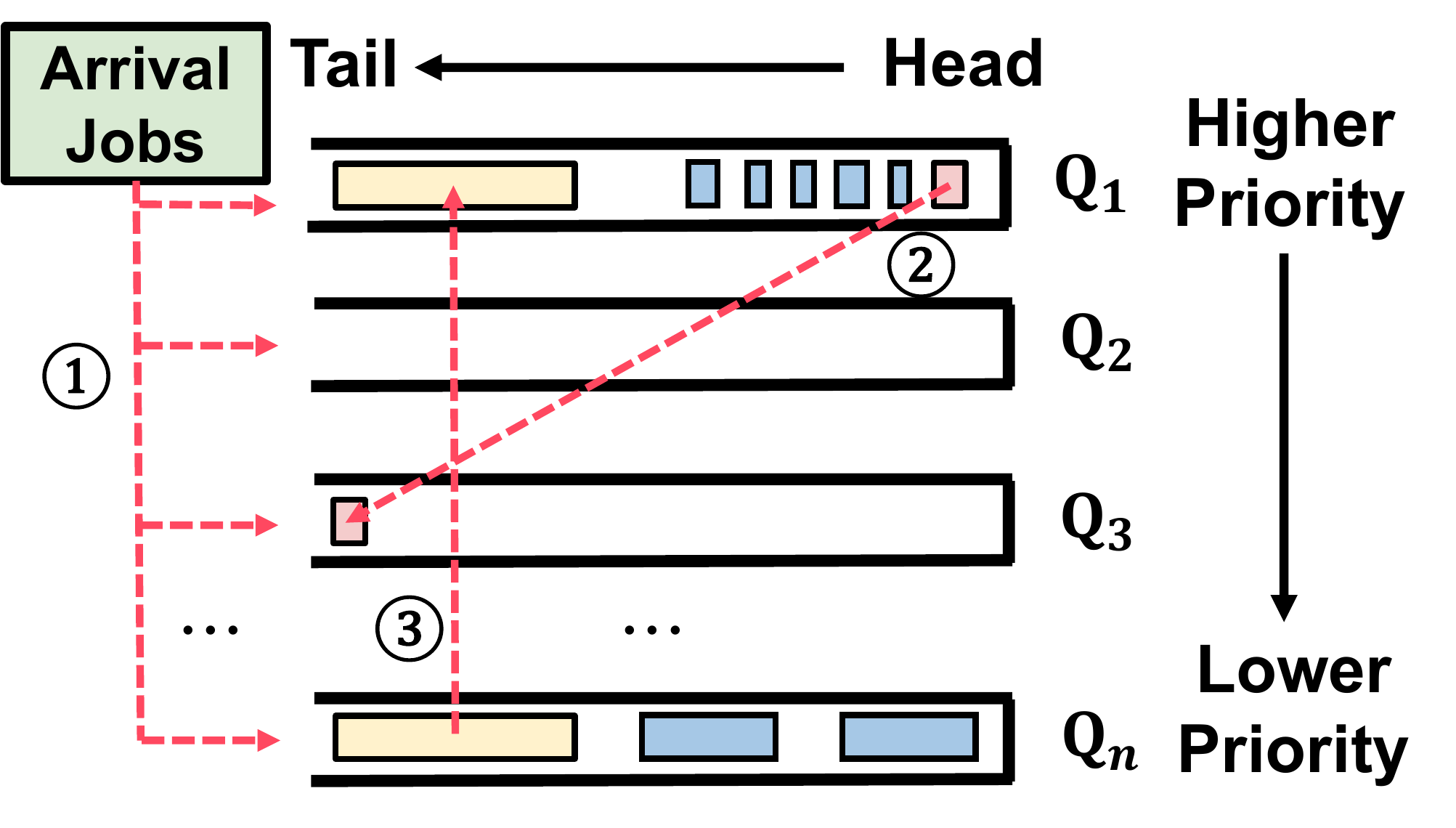}
    \vspace{-0.1in}
    \caption{Skip-join MLFQ with starvation prevention.}
    \vspace{-0.15in}
    \label{fig:design:priority_queue}
\end{figure}

\parabf{Our solution: skip-join MLFQ.}
The key insight of our design is to leverage the \emph{semi}
information-agnostic setting of LLM inference
to address the aforementioned issues of the strawman solutions. 
While the number of iterations (i.e., the output length) remains unknown beforehand, the execution time of each iteration is predictable.
\revisionatc{For each iteration, the execution is similar to the traditional one-shot DNN inference, whose execution time is highly predictable~\cite{gujarati2020serving,li2023alpaserve}. A lightweight profiling process can easily collect accurate iteration time under different hardware, model specifications, and input lengths in advance.}
Revisiting Figure~\ref{fig:design:iteration_time}, the
initialization phase time exhibits a positive correlation with the input length while fixing the hardware
and model. As for the decoding phase, the iteration time is roughly constant.

Based on this insight, we propose a skip-join MLFQ scheduler tailored for LLM
inference. Our scheduler efficiently manages the movement of jobs among various
priority queues in a skip-join manner. In the MLFQ, we have $n$ priority queues,
namely $Q_1, Q_2, ..., Q_n$, each with a distinct quantum $q_1 < q_2 < ... <
q_n$. The conventional MLFQ scheduler initially assigns a newly arrived job to
the highest priority queue, i.e., $Q_1$. Once the job exhausts the allocated
quantum in $Q_1$, it is subsequently demoted to $Q_2$. As shown in
Figure~\ref{fig:design:priority_queue}, \sysname differs from the original MLFQ
in that when a job arrives, \sysname leverages accurate profiling to predict the
initialization phase time ($t_{init}$) and \ding{182} skip-joins the job to the
highest priority queue ($q_i$) subject to $q_i \geq t_{init}$. When a job
consumes its allotted quantum before completion, the scheduler demotes
\ding{183} the job based on its current priority and next iteration time.

\parabf{Avoiding perpetual starvation.}
It is important to note that the skip-join and demotion operations may result in
starvation for jobs with long input and output. To address this problem, the
scheduler periodically examines the starved jobs and \ding{184} promotes them to
the highest priority queue, i.e., $Q_1$. This allows \sysname to address
head-of-line blocking while mitigating starvation. We evaluate tail latency
and show the effectiveness of the starvation prevention mechanism in~\S\ref{sec:evaluation:end}.

\begin{algorithm}[t]
    \small
    \caption{\small Skip-Join Multi-Level Feedback Queue Scheduler}
    \label{algorithm:skip_join}
    \begin{algorithmic}[1]
    \State {\bf Input:} Queues $Q_{1}, Q_{2}, ..., Q_{n}$, newly arrived jobs $J_{new}$
    \State {\bf Output:} Jobs to be executed $J_{out}$ for one iteration
    \Procedure{SkipJoinMLFQScheduler}{}
    \State Initialization: $J_{out} \gets \emptyset$.
    \State \textcolor{ao}{// Skip-join newly arrival jobs.}
    \For{$job \in J_{in}$}
        \State $init\_time \gets P.\textcolor{method}{getNextIterTime}(job)$
        \State $p_{job} \gets \textcolor{method}{min\ i,\ s.t.\ q_i \geq init\_time}$
        \State $Q_{p_{job}}.\textcolor{method}{push}(job)$
    \EndFor
    \For{$job \in \{Q_{1}, Q_{2}, ..., Q_{n}\}$}
        \State $job.\textcolor{method}{outputNewGeneratedToken}()$
        \State $p_{job} \gets job.\textcolor{method}{getCurrentPriority}()$
        \If{$job.\textcolor{method}{isFinished}()$}
            \State $Q_{p_{job}}.\textcolor{method}{pop}(job)$
        \EndIf
        \State \textcolor{ao}{// Demote jobs.}
        \If{$job.\textcolor{method}{depleteQuantum}()$}
            \State $Q_{p_{job}}.\textcolor{method}{pop}(job)$, $Q_{p_{job}+\eta}.\textcolor{method}{push}(job)$
        \EndIf
        \State \textcolor{ao}{// Promote starved jobs.}
        \If{$job.starveTime \geq \alpha$}
            \State $Q_{p_{job}}.\textcolor{method}{pop}(job)$, $Q_{1}.\textcolor{method}{push}(job)$
            \State $job.starveTime \gets 0$
        \EndIf
    \EndFor
    \State \textcolor{ao}{// Schedule jobs to execute.}
    \For{$job \in \{Q_{1}, Q_{2}, ..., Q_{n}\}$}
        \If{$job.\textcolor{method}{isReady}()$ \textbf{and} $|J_{out}| < MaxBatchSize$}
            \State $J_{out}.\textcolor{method}{push}(job)$
        \EndIf
    \EndFor
    \EndProcedure
    \end{algorithmic}
\end{algorithm}

\begin{figure}[t]
    \centering
    \includegraphics[width=\linewidth]{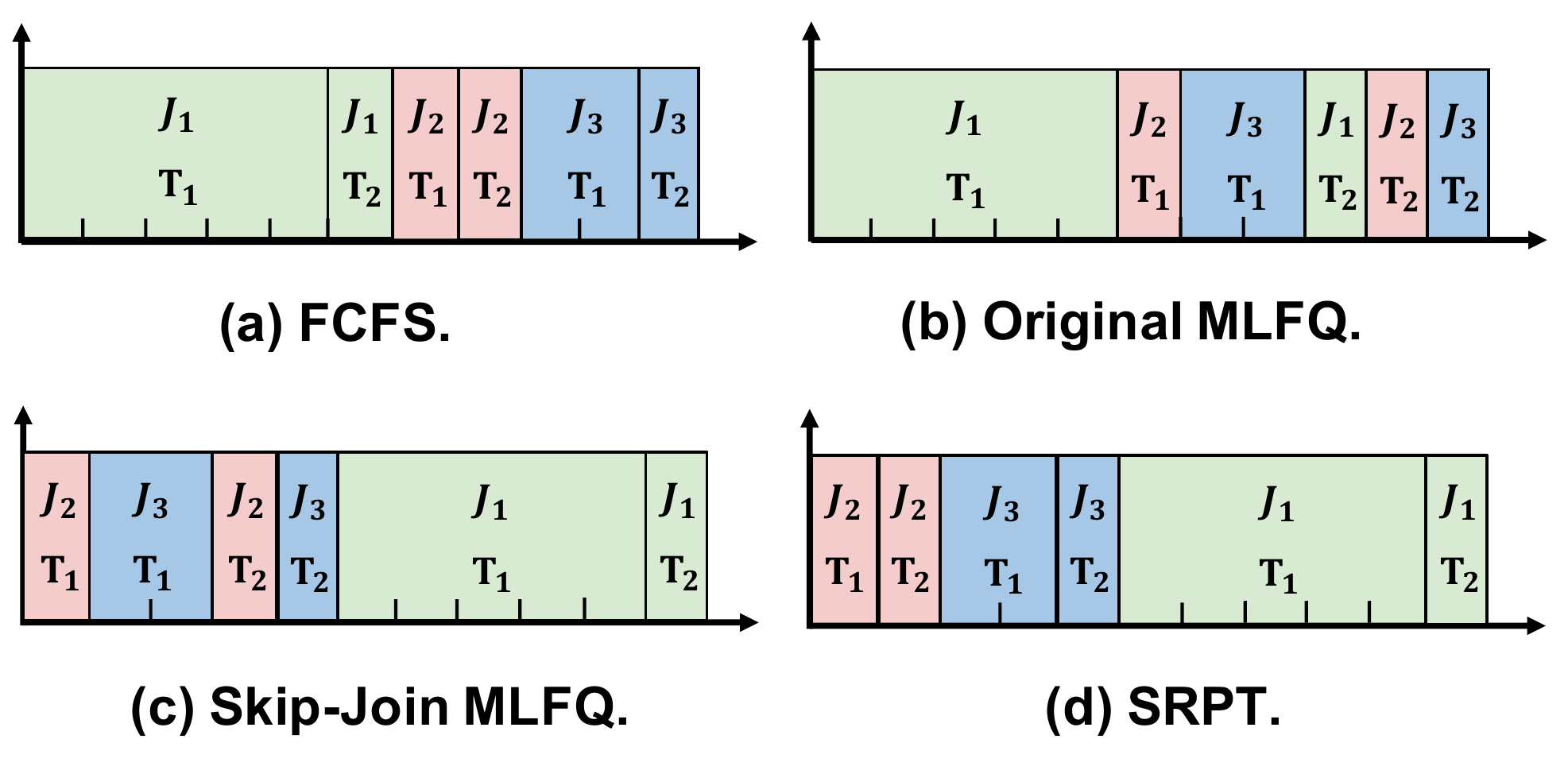}
    \vspace{-0.2in}
    \caption{Execution timeline under different scheduling algorithms.}
    \vspace{-0.15in}
    \label{fig:design:example}
\end{figure}

\parabf{Example.}
Figure~\ref{fig:design:example} shows an example to demonstrate the
effectiveness of \sysname's skip-join MLFQ scheduler.
In the example, three jobs arrive at the same time in the order of
$J_{1}, J_{2}, J_{3}$. $T_1(J_i)$ denotes the initialization phase
time of job $J_i$, and $T_2(J_i)$ denotes the decoding phase time.
We assume that both skip-join and original MLFQ utilize four priority queues
with quantum values of 1, 2, 4, and 8. Additionally, SRPT serves as the oracle
with the optimal average latency.

As Figure~\ref{fig:design:example} shows, the average latency of FCFS, original MLFQ, skip-join MLFQ, and
SRPT are 4.23, 5, 3.3, and 3, respectively. FCFS and original MLFQ encounter the head-of-line
blocking problem, where job $J_1$ blocks the remaining jobs, leading to long average latency.
Skip-join MLFQ addresses this issue by skip-joining job $J_1$ to the
low-priority queue, achieving performance similar to optimal SRPT.
Generally, algorithms that have access to more information perform
better than those with limited information.

\parabf{Algorithm.} 
Algorithm~\ref{algorithm:skip_join} shows the pseudo-code of the skip-join MLFQ
scheduler. The scheduler has a set of priority queues $Q_{1}, Q_{2}, ..., Q_{n}$
with quantum values $q_{1}, q_{2}, ..., q_{n}$, and receives a set
of newly arrived jobs $J_{new}$. It schedules a batch of $MaxBatchSize$ jobs for
execution. The skip-join part (\ding{182} in
Figure~\ref{fig:design:priority_queue}) corresponds to lines 6--9, and the
demotion and starvation prevention parts (\ding{183} and \ding{184} in
Figure~\ref{fig:design:priority_queue}) correspond to lines 16--17 and lines
19--21, respectively.
There are two notable details. First, the scheduler demotes a job to an $\eta$ times lower priority queue based on its next iteration time.
\sysname sets the quantum of the lower priority queue to two times of that of the higher priority queue, \revisionatc{which aligns with previous work~\cite{gu2019tiresias} on MLFQ}.
The quantum of the highest priority queue is set to the minimum iteration time.
The second detail concerns how the scheduler identifies starved jobs
governed by the parameter $\alpha$.
\revisionatc{\sysname tunes $\alpha$ based on the user-specified SLO, which is set to 300 ms by default.}

\subsection{Proactive Key-Value Cache Management}
\label{subsec:kvcache}

\begin{figure}[t]
    \centering
    \includegraphics[width=0.9\linewidth]{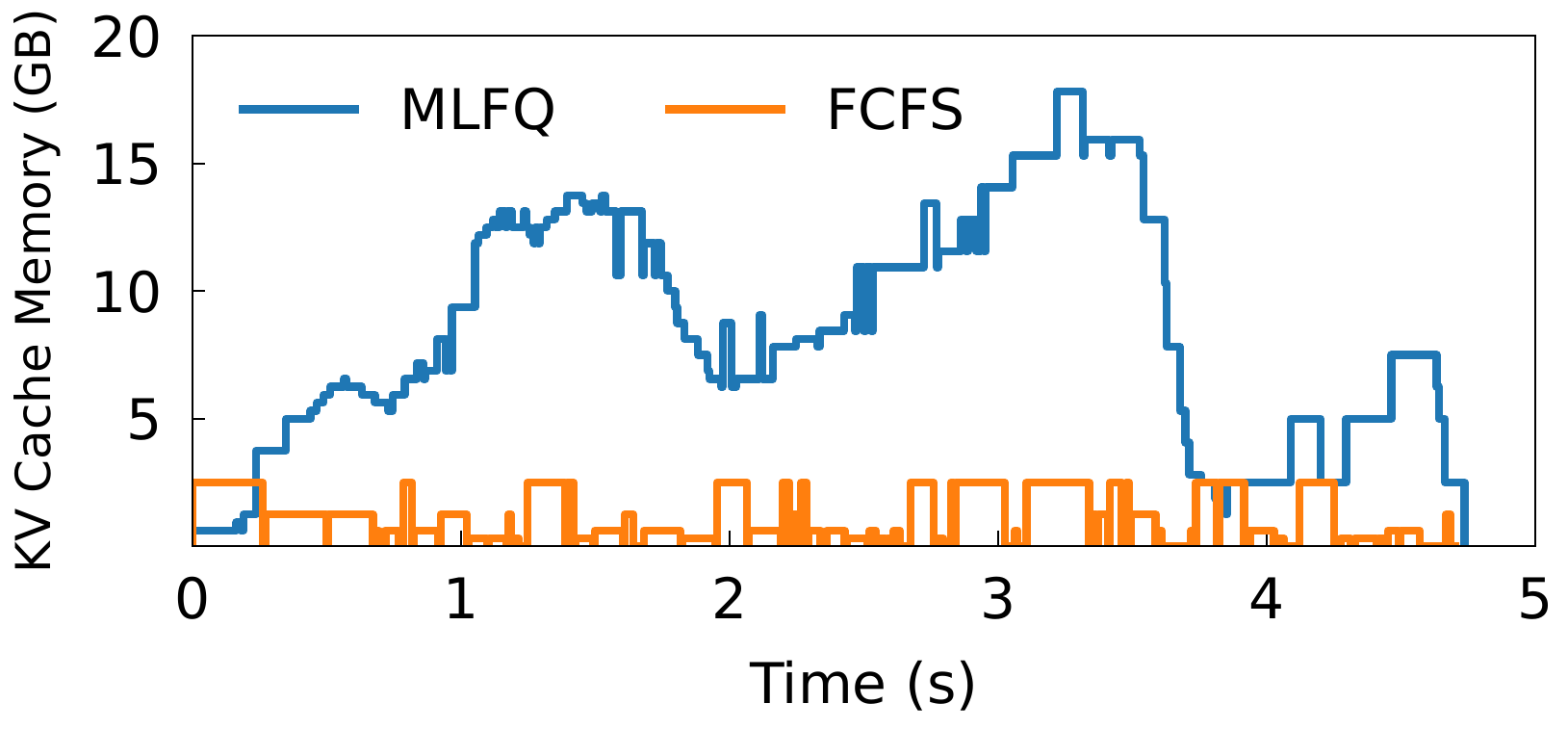}
    \vspace{-0.1in}
    \caption{The key-value cache memory consumption for OPT 2.7B under different schedulers. The workload follows a Gamma Process with rate=64 and CV=4. The maximum output length is set to 20..}
    \vspace{-0.1in}
    \label{fig:design:memory}
\end{figure}

Although the skip-join MLFQ scheduler provides iteration-level preemption to approximate SRPT to achieve lower latency without prior knowledge of the exact job size, it exacerbates the pressure of GPU memory consumption.
Figure~\ref{fig:design:memory} shows the key-value
cache memory consumption of FCFS and skip-join MLFQ for OPT 2.7B model under a synthetic workload.
Although we choose a relatively small model and limit the maximum output length to 20, the peak KV cache memory overhead for skip-join MLFQ can be $7\times$ larger than that of FCFS.
The GPU memory demand becomes even more pronounced when deploying larger LLMs like OPT 175B.

The reason under the hood is that compared to the run-to-completion policy in the existing serving systems, iteration-level preemption provided by the skip-join MFLQ increases the number of ongoing jobs in the system.
Except for the key-value tensors of running jobs, the skip-join scheduler also needs to store the key-value tensors for preempted jobs at the pending state.
Unlike process states in the traditional operating system schedulers, the intermediate state, i.e., key-value tensors, of each job is much larger.
Formally, for a particular LLM inference serving job,
denote the input sequence length by $s$,
the output sequence length by $t$,
the hidden dimension of the transformer by $h$,
and the number of transformer layers by $l$.
If the model weights and all computations are in FP16,
the total number of bytes to store the key-value cache for this single job is $4 \times lh(s+t)$.
Take OPT 175B as an example $(l=96, h=12288)$.
Given an input sequence length $s=512$ and a minimum output sequence length $t=1$,
the GPU memory overhead for a single job is as high as $2.3$GB. As the generation continues,
its output sequence length $t$ will increase,
which further increases the GPU memory overhead.

At the same time, GPU memory is a scarce resource when deploying LLMs.
Typically, GPU memory is much smaller than the host memory.
For instance, NVIDIA A100 GPU has a maximum of 80 GB GPU memory.
Besides, a large portion of GPU memory is provisioned to store weights of LLMs.
The space to store key-value tensors for jobs is limited.
As a result, the GPU memory capacity constrains the potential benefits of the skip-join MLFQ scheduler.

\parabf{Strawman solution 1: defer newly arrived jobs.}
To avoid out-of-memory (OOM) errors,
a naive solution is to simply \emph{defer} the execution of newly arrived jobs when the GPU memory is not sufficient and keep scheduling current in-memory jobs until they finish.
This straightforward solution is widely used in existing serving systems, such as vLLM~\cite{vllm}.
In this manner, although new jobs are assigned with higher priority,
they are blocked to await the free memory space.
Under extreme GPU memory-constrained settings (e.g., long sequence inference),
this solution would degenerate MLFQ to FCFS, which again suffers from head-of-line blocking.

\parabf{Strawman solution 2: kill and re-compute low-priority jobs.}
Another straightforward solution is to \textit{kill} some low-priority jobs and
release their key-value cache to make room for newly arrived high-priority jobs.
This solution has two problems. First, the killed jobs lose their generation
states, necessitating to rebuild their key-value tensors.
This results in the waste of valuable computational resources and time.
Second, it may cause deadlocks.
When the high-priority jobs arrive, ongoing jobs with lower priority
are killed.
Due to the starvation avoidance,
the killed jobs may be promoted to the highest-priority queue if they wait longer than the specified $STARVE_LIMIT$.
In this case, the promoted job may kill the currently executing job, which may just kill the promoted job in the previous step.
It potentially results in a deadlock.

\parabf{Our solution: proactive key-value cache swapping.}
Under the strict GPU memory capacity constraints,
the two strawman solutions have to sacrifice either the performance of newly arrived jobs or the efficiency of low-priority jobs.
To overcome this dilemma,
our key observation is that the key-value tensors only need to be reserved in the GPU memory when their corresponding jobs get scheduled.
Based on this observation,
\sysname extends the space of the key-value cache from GPU memory to the host memory.
\sysname swaps out inactive key-value tensors of jobs to the host memory to accommodate additional pending jobs,
and swaps in key-value tensors back to the GPU memory for upcoming jobs.

However,
the overhead of swapping is not negligible compared to the token generation time.
When deploying OPT 175B on 16 NVIDIA A100 GPUs,
the key-value tensors of a job can occupy 2.3 GB memory.
The token generation time in the decoding phase is about 60 ms,
while the time to swap the key-value tensors between host memory and GPU memory with PCIe 4.0$\times$16 full bandwidth is about 36 ms.
Therefore, a simple reactive swapping mechanism that processes swapping and inference sequentially introduces a large overhead.

\begin{figure}[t]
    \centering
    \includegraphics[width=\linewidth]{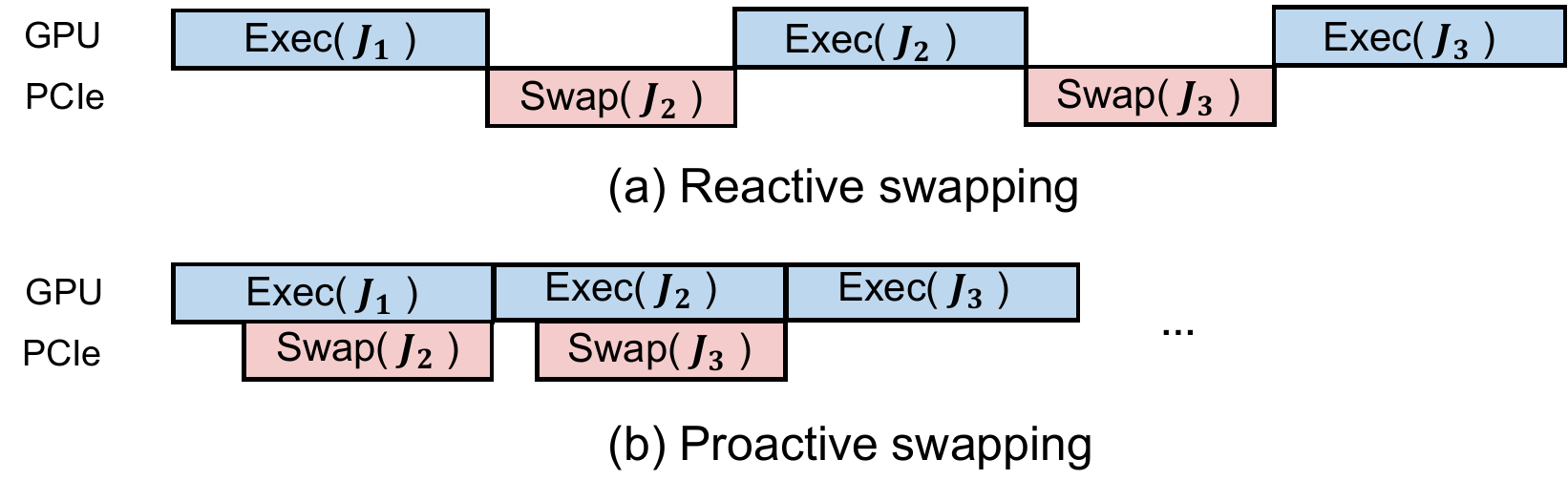}
    \vspace{-0.3in}
    \caption{\revisionatc{Comparison of reactive and proactive swapping.}}
    \vspace{-0.1in}
    \label{fig:design:swap}
\end{figure}

Instead, \sysname employs a \emph{proactive} key-value cache swapping algorithm to mitigate the adverse effects of swapping overhead.
The key insight is to overlap the LLM inference for running jobs with the data transmission for pending jobs so that the swapping overhead is out of the critical path of LLM inference.
\revisionatc{Figure~\ref{fig:design:swap} illustrates an example.
Instead of swapping key-value tensors of pending job $J_2$ after job $J_1$ is preempted or finished, the proactive algorithm swaps the key-value tensors of $J_2$ in advance.
In this way, the swapping overhead of $J_2$ effectively overlaps with the GPU kernel execution of $J_1$, thereby achieving high GPU utilization.
As swapping in one job consumes expensive GPU memory, the job swapping order is crucial for GPU memory efficiency.}

\parabf{Job swapping order.}
Frequently swapping in and out unnecessary key-value tensors incurs additional
thrashing overhead if swapping in and out one job with high priority.
The swapping overhead can increase to exceed the execution time,
leading to a deterioration in the performance of overlapping.
To address this issue,
\sysname calculates the estimated next scheduled time (ENST) for each job to decide the swapping order.
The ENST is the time when the job will be scheduled to execute next time.
The job with the largest ENST will be swapped out first,
and the job with the smallest ENST will be swapped in first.
Typically, a job with lower priority is scheduled for later execution.
However, owing to the starvation prevention mechanism,
a job of lower priority might be elevated to a higher priority queue.
Consequently, even a low-priority job can sometimes be executed first.

In this case, for job $i$, \sysname considers the time to promote
this job and the sum of execution time of all jobs with higher priorities before
executing $i$ simultaneously. Formally, let the time threshold for promoting job $i$ be $T_{promote}(i)$. As for
the sum of execution time of all jobs with higher priorities before executing $i$,
we assume those jobs do not finish earlier before being demoted to
the priority queue of job $i$. The execution time of job $j$ with a higher
priority can be calculated as follows (i.e., job $j$ is demoted from $j.priority$ to $i.priority$ one by one):
\[
    T_{execute}(i, j) = \sum_{i.priority < k \leq j.priority} q_k
\]
where $i.priority$ is the priority of job $i$, and $q_k$ is the quantum of the priority queue with priority $k$.
Based on this, the sum of execution time of all jobs with higher priorities than job $i$ is defined as:
\[
    T_{execute}(i) = \frac{1}{B}\sum_{i.priority < j.priority} T_{execute}(i, j)
\]
where $B$ is the maximum batch size of jobs.
At last, taking both the promotion for starvation prevention
and the execution of higher priority jobs into consideration, the ENST of job $i$ is calculated as:
\[
    ENST(i) = \min(T_{promote}(i), T_{execute}(i))
\]

This ENST definition serves as a means to estimate the expected scheduling time for the next generation of job $i$.
Therefore, using this metric to decide the order of swapping makes the key-value tensors of active jobs reside predominantly in GPU
memory, and those of inactive jobs are more inclined to reside in host memory.

\parabf{Handling a burst of new jobs.}
The proactive key-value cache swapping strategy is designed for
the skip-join MLFQ scheduler. In scenarios where a
significant influx of new jobs (with high priority) occurs, the cache management system
is forced to evict jobs reactively, adversely affecting the performance of these new
jobs. To mitigate this, \sysname reserves
some idle key-value cache slots specifically for new jobs, ensuring immediate
availability without the need for reactive job swapping. This approach guarantees
the performance of new jobs. \revisioneurosys{The number of idle slots is based
on historical job arrival patterns. A higher frequency of
job bursts necessitates a larger number of reserved slots.}

\subsection{Support for Distributed LLM Serving}
\label{subsec:design:distributed_support}

\begin{figure}[t]
    \centering
    \includegraphics[width=0.8\linewidth]{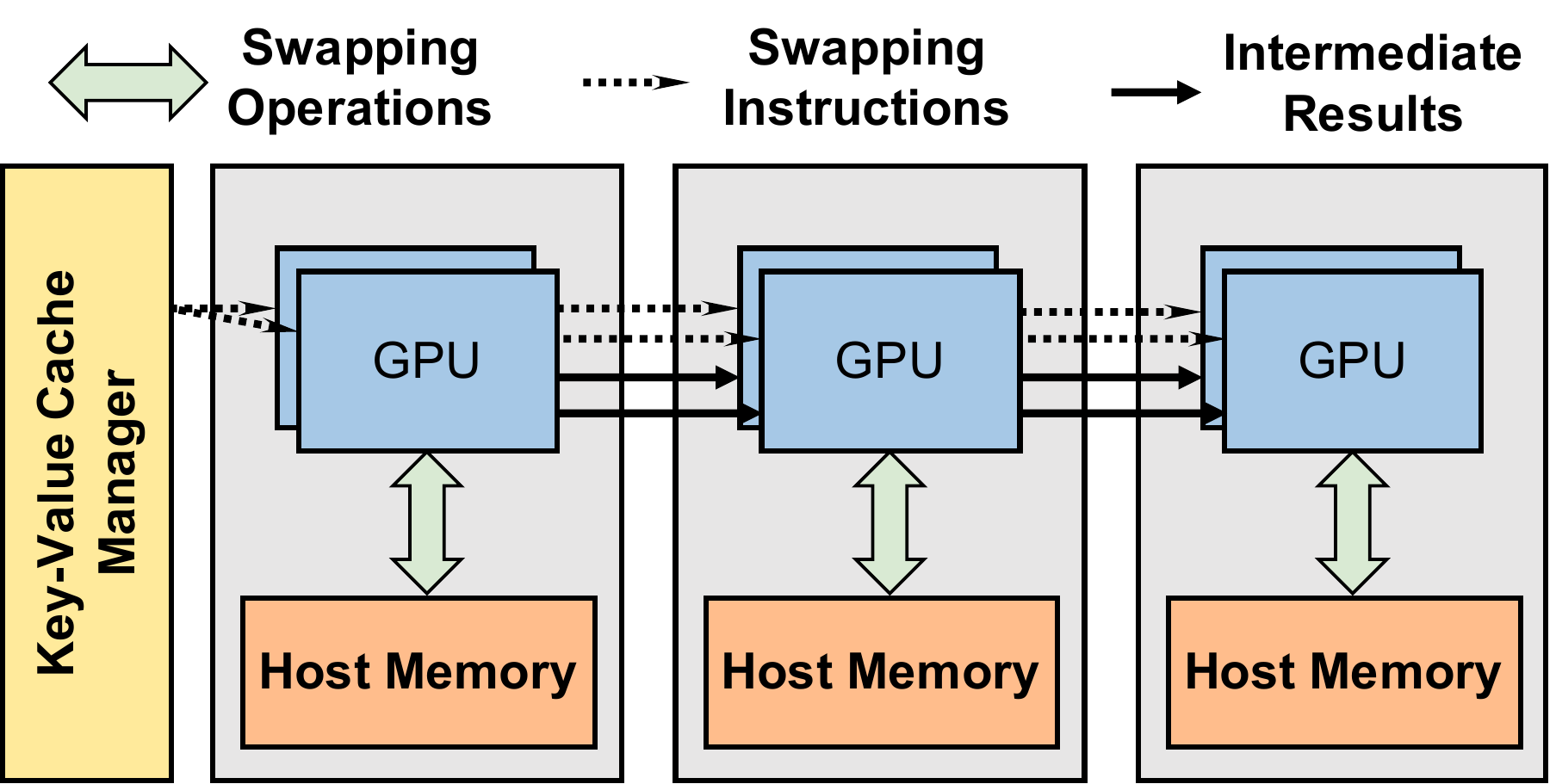}
    \vspace{-0.03in}
    \caption{Overlapping key-value cache offloading with intermediate result transmission.}
    \vspace{-0.1in}
    \label{fig:design:distributed_kv_cache}
\end{figure}

Previous research shows that the effectiveness of LLMs empirically adheres to
the scaling law concerning the quantity of model
parameters~\cite{kaplan2020scaling}. However, it is important to note that the
memory usage of an LLM also exhibits proportionality to the number of
parameters. A prime example is OPT 175B, which, even when stored in
half-precision, demands a staggering 350GB of GPU memory solely to accommodate
its weights. Furthermore, additional memory is required for handling
intermediate states during runtime. Therefore, LLM often needs to be split into
multiple pieces and served in a distributed manner with multiple GPUs.

Tensor parallelism~\cite{shoeybi2020megatronlm, narayanan2021efficient} and pipeline parallelism~\cite{huang2019gpipe, pipedream}
are two most widely-used techniques for distributed LLM serving.
\sysname supports the hybrid of
these two parallel techniques for serving LLMs. An LLM is composed of a series
of operators over multi-dimensional tensors. Tensor parallelism splits each
operator across multiple devices, with each device executing a portion of the
computation in parallel. Additional communication overhead is required to split
the input and gather the output from different GPUs. Tensor parallelism
substantially augments both computational and memory resources available to
a single job, consequently reducing the time of each iteration.

Pipeline parallelism splits the entire operators of an LLM computation graph into
multiple stages and executes them on different GPUs in a pipeline fashion.
During inference, each stage computes a part of the entire computation graph and
transmits the intermediate results to the next stage in parallel. Pipeline
parallelism requires less communication overhead compared to tensor parallelism,
while also affording LLMs the ability to surpass the memory constraint of an individual GPU.
Since multiple processing batches are under processing simultaneously in different
stages, \sysname needs to handle multiple batches in the distributed engine at
the same time.

\parabf{Job scheduling in distributed serving.}
In the traditional MLFQ, if no new job arrives, the scheduler
schedules the job with the highest priority and executes it until it finishes or
is demoted. However, with pipeline parallelism, the scheduler schedules at
the granularity of individual stage. Once a job completes its first stage
and transmits the intermediate results to the subsequent stage,
a decision point arises for the scheduler regarding the next job to set in motion.
In this case, the scheduler cannot follow the
traditional MLFQ that keeps scheduling the same job until demotion, because the job is still in progress.
To preserve the semantics of MLFQ, \sysname still keeps the
running job in the priority queue, but schedules the highest priority
job in the pending state. Thus, the early jobs in a queue can expedite their quantum completion.

\parabf{Key-value cache management in distributed serving.}
Given that the key-value cache occupies a large fraction of GPU memory, the
key-value cache of \sysname is also partitioned across multiple GPUs. 
In LLM inference, each key-value tensor is used by the same
stage of the LLM. Therefore, \sysname partitions key-value tensors as tensor
parallelism requires, and assigns each key-value tensor to the corresponding GPU
so that all computation on a GPU only needs local key-value tensors on the same GPU.

The proactive key-value cache swapping mechanism of \sysname is also
distributed. Because different stages of the LLM process different jobs at the
same time, each stage may offload or upload different key-value tensors
independently. To reduce redundant control, before processing the intermediate
result sent from the previous stage, the current stage does the same offloading
or uploading action as the previous stage does. The intermediate result
transmission and key-value cache swapping occur in parallel, so the overhead of
key-value cache swapping is further reduced. As shown in
Figure~\ref{fig:design:distributed_kv_cache}, when the intermediate result is
sent to the next stage, the next stage receives the swapping instructions and
can swap the key-value cache at the same time if needed. The key-value cache
swapping mechanism only needs to decide the offloading or uploading of the first
stage. When using tensor parallelism splitting the first stage into multiple
chunks, a centralized key-value cache swapping manager instructs all chunks in
the first stage to offload or upload the key-value tensors owned by the same
job.
\section{Implementation}
\label{sec:implementation}

\sysname is a distributed LLM inference serving system with a RESTful API frontend, a scheduler, and a distributed execution engine. The frontend and scheduler are implemented with 2.9K lines of Python code. The distributed execution engine is implemented with 8.1K lines of C++/CUDA code. The frontend supports OpenAI API compatible interface where clients can specify the sampling parameters like maximum output length and temperature. The scheduler implements the skip-join MLFQ and proactive swapping policies. The distributed execution engine uses Ray~\cite{ray} actor to implement GPU workers which execute the LLM inference and manage the key-value cache in a distributed manner. We implement popular open-source LLMs such as OPT in C++ to achieve better performance and scalability than the popular Python implementations in Huggingface~\cite{wolf2020huggingfaces}. We also implement custom CUDA kernels to support Orca's~\cite{yu2022orca} iteration-level scheduling and vLLM's~\cite{vllm} PagedAttention.

\section{Evaluation}
\label{sec:evaluation}

\begin{table}[t]
    \centering
    \resizebox{1\linewidth}{!} {
    \begin{tabular}{lrccc}
        \toprule
        \textbf{Model} & \textbf{Size} & \textbf{\# of Layers} &
        \textbf{\# of Heads} & \textbf{Hidden Size} \\
        \midrule
        OPT-13B & 26GB & 40 & 40 & 5120 \\
        OPT-66B & 132GB & 64  & 72 & 9216 \\
        OPT-175B & 350GB & 96 & 96 & 12288 \\
        \bottomrule
    \end{tabular}
    }
    \caption{Model configurations.}
    \vspace{-0.1in}
    \label{tab:evaluation:models}
\end{table}

\begin{table}[t]
    \centering
    \resizebox{1\linewidth}{!} {
    \begin{tabular}{lcccc}
        \toprule
        & \textbf{FasterTransformer~\cite{fastertransformer}} & \textbf{vLLM~\cite{vllm}} &
        \textbf{\sysname-FCFS} & \textbf{\sysname} \\
        \midrule
        IP &            &            &            & \checkmark \\
        IS &            & \checkmark & \checkmark & \checkmark \\
        PA &            & \checkmark & \checkmark & \checkmark \\
        PP & \checkmark &            & \checkmark & \checkmark \\
        \bottomrule
    \end{tabular}
    }
    \vspace{-0.05in}
    \caption{\revisionatc{Comparison between \sysname and baselines. IP=Iteration-level Preemption, IS=Iteration-level Scheduling, PA=PagedAttention, PP=Pipeline Parallelism.}}
    \vspace{-0.1in}
    \label{tab:evaluation:baselines}
\end{table}

\begin{figure*}[t]
    \centering
    \includegraphics[width=\linewidth]{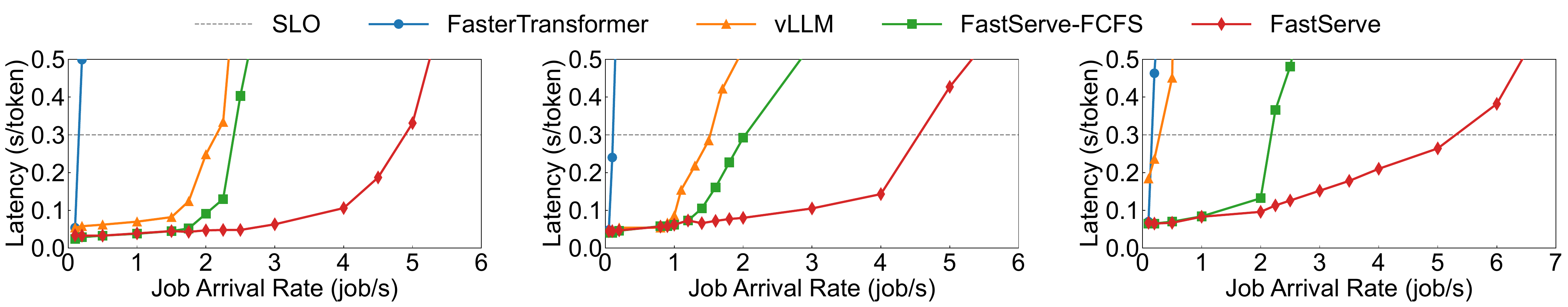}
    \hspace*{0.in}{(a) OPT-13B, 1 GPU, ShareGPT.}\hspace*{\dimexpr\linewidth/16\relax}{(b) OPT-66B, 4 GPUs, ShareGPT.}\hspace*{\dimexpr\linewidth/16\relax}{(c) OPT-175B, 16 GPUs, ShareGPT.}
    \vspace{0.1in}
    \includegraphics[width=\linewidth]{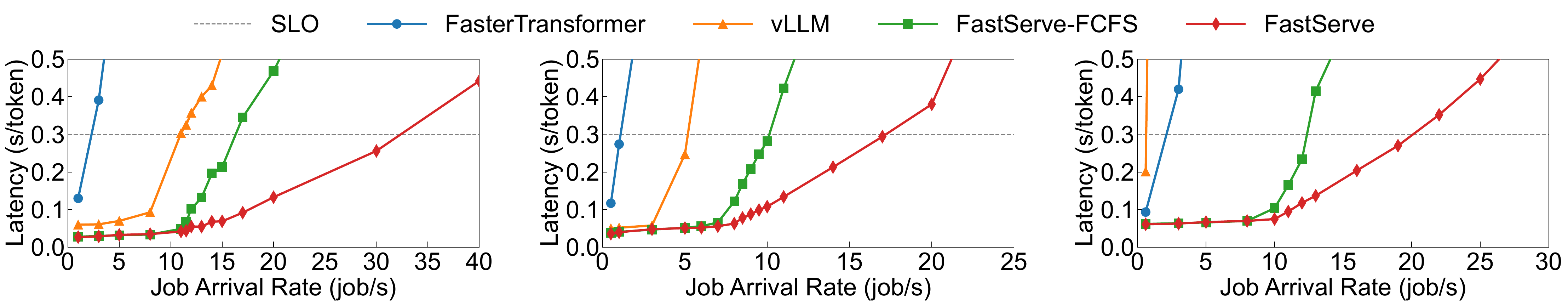}
    \hspace*{0.0in}{(a) OPT-13B, 1 GPU, Alpaca.}\hspace*{\dimexpr\linewidth/11\relax}{(b) OPT-66B, 4 GPUs, Alpaca.}\hspace*{\dimexpr\linewidth/11\relax}{(c) OPT-175B, 16 GPUs, Alpaca.}
    \vspace{-0.05in}
    \caption{Average latency of different serving systems with OPT models on real workloads.}
    \vspace{-0.2in}
    \label{fig:evaluation:end2end}
\end{figure*}

In this section, we first demonstrate the end-to-end performance improvements of
\sysname over state-of-the-art LLM serving systems. Then, we evaluate the design
choices \sysname and show the effectiveness of each component.

\subsection{Methodology}
\label{sec:evaluation:methodology}

\paraf{Testbed.} The end-to-end experiments (\S\ref{sec:evaluation:end}) use two
AWS EC2 p4d.24xlarge instances. Each instance is configured with eight NVIDIA
A100 40GB GPUs connected over NVLink, 1152 GB host memory, and PCIe
4.0$\times$16. Due to the limited budget, the experiments for design choices
(\S\ref{sec:evaluation:ablation}) use one NVIDIA A100 40GB GPU in our own
testbed to validate the effectiveness of each component.

\parabf{LLM models.} We choose the representative LLM family, OPT~\cite{zhang2022opt},
which is widely used in both academia and industry. We select common model sizes.
Table~\ref{tab:evaluation:models} lists the model configurations. We use FP16 precision in all experiments.

\parabf{Workloads.} Similar to prior work on LLM serving~\cite{vllm}, we
generate workloads based on ShareGPT~\cite{sharegpt} and Alpaca~\cite{alpaca} datasets. These datasets contain real-world inputs and outputs of LLM services.
The ShareGPT dataset is composed of user-shared conversations with ChatGPT~\cite{sharegpt}.
The Alpaca dataset is generated by GPT-3.5 with self-instruct~\cite{alpaca}.
Since these datasets do not include the arrival time,
we follow prior work~\cite{vllm} to generate the arrival time for each request following a Poisson process
parameterized by the arrival rate. 

\vspace{-0.1in}

\parabf{Evaluation metrics.} The user-perceived latency is a critical
measurement for interactive applications like ChatGPT. Specifically, similar to
prior work on LLM serving~\cite{yu2022orca, vllm}, average per-token latency is calculated
as the mean of every job's end-to-end latency divided by its output length. \revisionatc{In addition, we also report the P95 tail latency.}

For comparison, we set a latency SLO and compare the maximum
throughput each system can achieve under the SLO. We follow prior
work~\cite{zygos} to set the latency SLO to 10$\times$ of the latency of a single iteration in the decoding phase. Specifically, we set SLO to 0.3 seconds based on our profiling.

\vspace{-0.1in}

\parabf{Baselines.} We compare \sysname with three baselines.
For fair comparison, all baselines use the same tensor parallelism size,
pipeline parallelism size, and batch size as \sysname, except that vLLM only uses tensor parallelism to serve OPT-175B, because it does not support pipeline parallelism. \revisionatc{Table~\ref{tab:evaluation:baselines} shows the comparison between \sysname and baselines.}
\begin{itemize}[leftmargin=*]
    \item \textbf{FasterTransformer}~\cite{fastertransformer}: It is a production-grade inference engine from NVIDIA. It supports both tensor parallelism and pipeline parallelism. However, it adopts job-level scheduling and short jobs are blocked by long jobs in the same batch. \revisionatc{We use FasterTransformer v5.3.}

    \item \textbf{vLLM}~\cite{vllm}: It is the state-of-the-art LLM
    serving system that supports iteration-level scheduling~\cite{yu2022orca} and PagedAttention~\cite{vllm} to reduce memory fragmentation caused by key-value cache.
    However, it uses a simple
    FCFS scheduler with run-to-completion execution, which suffers from
    head-of-line blocking. \revisionatc{We use vLLM v0.1.7.}

    \item \textbf{\sysname-FCFS}: It uses the same distributed execution engine of \sysname, but it does not use techniques proposed in~\S\ref{sec:design}. This baseline helps differentiate the speedup brought by the techniques proposed in this paper from
    that by the efficient implementation of \sysname.
\end{itemize}

\subsection{End-to-End Performance}
\label{sec:evaluation:end}

We compare the end-to-end performance of \sysname to the
three baseline systems under ShareGPT and Alpaca workloads on OPT-13B, OPT-66B
and OPT-175B in Table~\ref{tab:evaluation:models}.

The first row of Figure~\ref{fig:evaluation:end2end} shows the end-to-end
performance of all the systems under the ShareGPT dataset. Although
FasterTransformer implements highly optimized GPU kernels for LLM inference, it
does not support iteration-level scheduling. It cannot return early finished
jobs in the ongoing batch and add new jobs into the batch to reduce latency. As
a result, FasterTransformer suffers from the significant head-of-line blocking
even when the job arrival rate is small. \sysname outperforms FasterTransformer
by 31.5--74.9$\times$ in terms of throughput under the SLO. As the
state-of-the-art serving system, vLLM is equipped with most of the techniques to
accelerate inference and reduce GPU memory consumption. However, because vLLM
schedules jobs in an FCFS manner, a large portion of the end-to-end latency is
the queuing delay. Optimizing the execution time of the LLM inference job is not
enough. Equipped with the skip-join MLFQ scheduler, \sysname can significantly
reduce the queuing delay and outperform vLLM by 2.3--18.3$\times$. It is worth
noting that \sysname-FCFS also outperforms vLLM, because it uses more efficient
C++ implementation and fuses more operations into fewer GPU kernels. But it
still suffers from the head-of-line blocking problem, which makes it slower than
\sysname by 2--4$\times$.

\begin{figure*}[t]
    \centering
    \includegraphics[width=\linewidth]{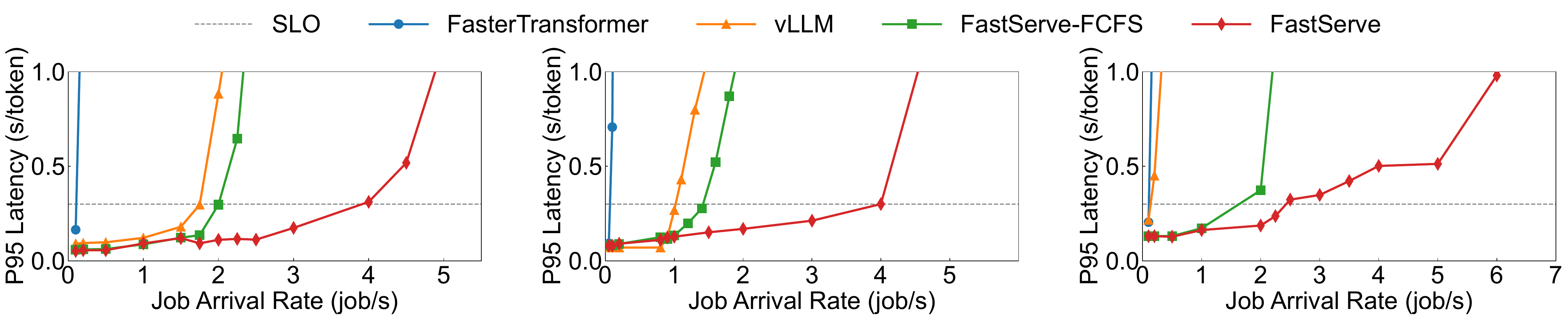}
    \hspace*{0.in}{(a) OPT-13B, 1 GPU, ShareGPT.}\hspace*{\dimexpr\linewidth/14\relax}{(b) OPT-66B, 4 GPUs, ShareGPT.}\hspace*{\dimexpr\linewidth/14\relax}{(c) OPT-175B, 16 GPUs, ShareGPT.}
    \vspace{-0.05in}
    \caption{Tail latency of different serving systems with OPT models on real workloads.}
    \vspace{-0.25in}
    \label{fig:evaluation:fairness}
\end{figure*}

\begin{figure}[t]
    \centering
    \includegraphics[width=\linewidth]{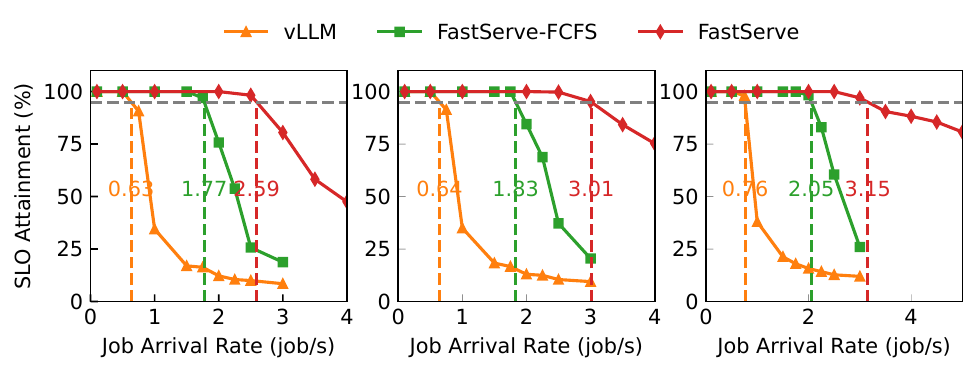}
    \hspace*{0.2in}{(a) 5$\times$ SLO.}\hspace*{\dimexpr\linewidth/12\relax}{(b) 10$\times$ SLO.}\hspace*{\dimexpr\linewidth/12\relax}{(c) 20$\times$ SLO.}
    \vspace{-0.1in}
    \caption{P95 goodput under different SLOs.}
    \label{fig:evaluation:slo}
    \vspace{-0.15in}
\end{figure}

The second row of Figure~\ref{fig:evaluation:end2end} shows the experiment
results under the Alapca dataset. Since the job size of Alpaca is smaller than
that of the ShareGPT dataset, all serving systems can maintain low latency even
when the rate is relatively higher than that under the ShareGPT dataset.
However, the performance gain of \sysname is similar. Without iteration-level
scheduling, FasterTransformer is the slowest system and \sysname outperforms it
by 9.5--15.8$\times$. vLLM achieves better performance than FasterTransformer,
but it still suffers from the head-of-line blocking problem. As a result,
\sysname outperforms vLLM by 3--31.4$\times$. With our efficient implementation,
\sysname-FCFS also outperforms vLLM, but it is still slower than \sysname by
1.6--2$\times$.

\parabf{Impact on tail latency.}
A potential concern of preemptive scheduling and MLFQ is that it can cause
starvation for long jobs and hurt tail latency. \sysname incorporates a
starvation prevention mechanism in its skip-join MLFQ scheduler
(\S\ref{subsec:mlfq}). To demonstrate the effectiveness of the starvation
prevention mechanism, we measure the 95\% latency of all the systems under the
ShareGPT dataset. As shown in Figure~\ref{fig:evaluation:fairness}, \sysname
significantly improves the throughput of LLM inference jobs under the same SLO
requirement for tail latency. For example, when serving OPT-175B, compared to
\sysname-FCFS, \sysname improves the throughput by up to 1.5$\times$. \sysname
also outperforms \sysname-FCFS by 2--2.8$\times$ when serving OPT-13B and OPT-66B.
\sysname achieves up to 17.9$\times$ and 59.8$\times$ performance improvement
compared to vLLM and FasterTransformer, respectively. The results show that
although \sysname is designed to reduce average latency, it can also
significantly reduce tail latency of LLM inference jobs. Prioritizing short jobs
with the skip-join MLFQ scheduler can effectively reduce the head-of-line
blocking problem and does not hurt the tail latency. Even for long jobs,
\sysname can still accelerate them by reducing their queuing delay. The
starvation prevention mechanism ensures that long jobs can be scheduled in a
reasonable time.

\parabf{Impact on goodput.}
\revisionnsdi{We further investigate the impact on goodput for different systems under various SLOs when serving OPT-13B model. Similar to previous works~\cite{li2023alpaserve, zhong2024distserve, wu2024loongserve}, we measure the P95 goodput, defined as the throughput when 95\% of jobs can be completed within the SLO of the initialization phase and the decoding phase. We set the SLO of two phases to 5$\times$, 10$\times$, and 20$\times$ of the latency in the corresponding phase under light load.}
\revisionnsdi{As shown in Figure~\ref{fig:evaluation:slo}, \sysname consistently achieves the highest P95 goodput across different SLOs. Specifically, \sysname outperforms vLLM by 4.1$\times$ to 4.7$\times$ and \sysname-FCFS by 1.46$\times$ to 1.64 $\times$. These results demonstrate that \sysname effectively improves the system's throughput without violating the SLOs of both the initialization phase and the decoding phase.}

\begin{figure}[t]
    \centering
    \includegraphics[width=\linewidth]{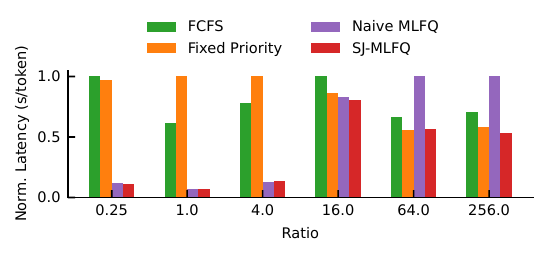}
    \vspace{-0.35in}
    \caption{Effectiveness of skip-join MLFQ.}
    \label{fig:evaluation:ablation_mlfq}
    \vspace{-0.15in}
\end{figure}

\subsection{Design Choices}
\label{sec:evaluation:ablation}
\paraf{Effectiveness of skip-join MLFQ.}
\revisionnsdi{To show the effectiveness of the skip-join MLFQ, we conduct a performance comparison against the FCFS, naive MLFQ, and Fixed Priority when serving OPT-13B. We use the ShareGPT dataset to generate jobs and alter the ratios between input and output lengths while preserving the original length distribution. This adjustment reflects the current industry trend of expanding input token limits in LLMs~\cite{wu2024loongserve,gemini,claude}.}
\revisionnsdi{Figure~\ref{fig:evaluation:ablation_mlfq} shows the results when varying the ratio between input and output lengths. The latency is normalized by the slowest system.
The FCFS consistently experiences high latency due to the head-of-line blocking issue, regardless of the ratio.
The naive MLFQ has good performance under a small ratio since the difference between the initialization and decoding phases is minimal.
However, as the ratio increases, the naive MLFQ struggles with the prolonged initialization phase.
In contrast, the Fixed Priority excels with larger ratios, where the initialization phase dominates
execution time, but underperforms when the ratio is low, as it neglects the decoding phase when setting priority.
Benefiting from a semi information-agnostic scheduling policy, the skip-join MLFQ consistently improves the performance compared to FCFS, naive MLFQ, and Fixed Priority by up to 8.9$\times$, 1.87$\times$, and 13.9$\times$.}

\begin{figure}[t]
    \centering
    \includegraphics[width=\linewidth]{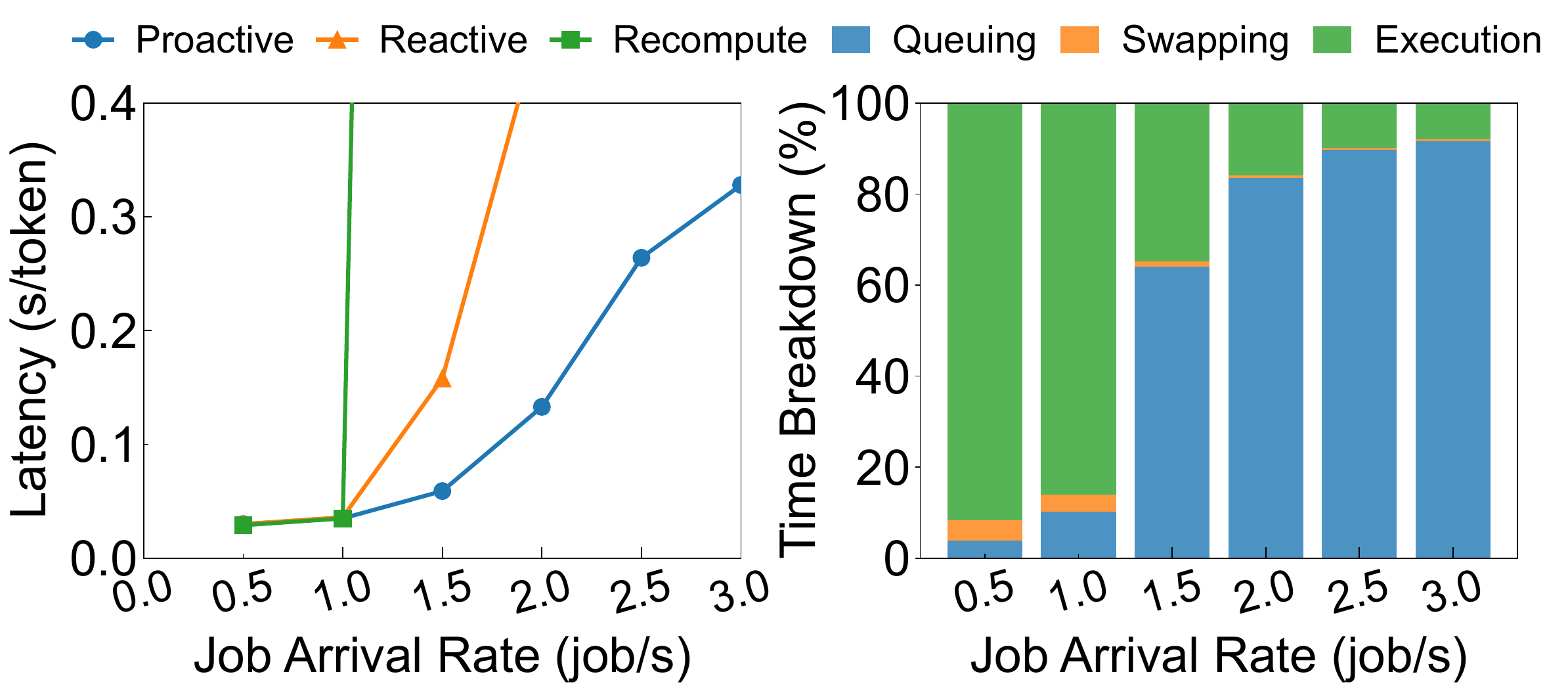}
    (a) OPT-13B on ShareGPT. \hspace*{0.2in} (b) Latency breakdown.
    \vspace{-0.1in}
    \caption{Effectiveness of proactive key-value cache management. (a) Comparison between different key-value cache management policy. (b) Latency breakdown of \sysname.}
    \vspace{-0.1in}
    \label{fig:evaluation:ablation_swapping}
\end{figure}

\parabf{Effectiveness of proactive key-value cache management.}
To show the effectiveness of the proactive key-value cache management, we evaluate the performance of \sysname with two baseline strategies Recompute and Reactive mentioned in \S\ref{subsec:kvcache} when serving OPT-13B on the ShareGPT dataset. Figure~\ref{fig:evaluation:ablation_swapping}\textcolor{green}{(a)} shows the results.
At low arrival rates, the GPU memory is sufficient to accommodate the key-value caches for all jobs, making the performance of the three strategies similar.
As the arrival rate increases, the GPU memory becomes insufficient so that systems have to preempt some key-value caches for other jobs, leading to distinct performance.

In such case, Recompute discards the key-value caches for low-priority jobs, increasing the re-computation overhead for these jobs' KV caches. As shown in Figure~\ref{fig:evaluation:ablation_swapping}\textcolor{green}{(a)}, this re-computation overhead makes
the proactive-swapping strategy outperforms recomputation by 2.7$\times$.

As for the Reactive, it swaps out low-priority jobs to host memory when GPU memory is inefficient and swaps in these jobs if needed.
The data transfer is in the critical path. Subsequent computation
must wait for these transfers. Conversely, proactive-swapping
anticipates the memory requirements of new incoming jobs and preempts
low-priority caches in advance. Similarly, when a high-priority job's cache
is detected in host memory and GPU memory is available, it will be
proactively swapped into the GPU memory. This allows the data transfer overlap with
computation and achieves 1.7$\times$ improvement over
reactive approach.

To further investigate the overhead of the proactive swapping mechanism, we split 
the end-to-end latency as three parts: queuing delay, execution time, and swapping time. 
The swapping time is the time when the job is blocked by the proactive swapping mechanism. 
As shown in Figure~\ref{fig:evaluation:ablation_swapping}\textcolor{green}{(b)}, the swapping time is less than 5\% of the end-to-end latency, which is negligible 
compared to the execution time and queuing delay. The reason confirms that the proactive swapping mechanism can overlap most of the swapping time with the execution time of other jobs. 
As a result, the proactive swapping mechanism nearly does not affect the end-to-end latency.

\section{Related Work}
\label{sec:related}

\paraf{Preemptive scheduling.} Many solutions for job scheduling in datacenters
use preemptive scheduling. Many networked systems~\cite{flowpreempt, alizadeh2013pfabric, chowdhury2014efficient, bai2015information} use preemptive flow scheduling to minimize flow completion time. Many schedulers for latency-sensitive datacenter workloads, such as Shinjuku~\cite{kaffes2019shinjuku}, Shenango~\cite{ousterhout2019shenango}, and Caladan~\cite{fried2020caladan}, also use fine-grained preemption and resource reallocation to optimize microsecond-scale tail latency. As for DL workloads, Tiresias~\cite{gu2019tiresias} uses MLFQ to optimize job completion time for distributed DL training jobs. Pipeswitch\cite{bai2020pipeswitch} and REEF\cite{preemptionInference} provide efficient GPU preemption to run both latency-critical and best-effort DL tasks on the same GPU. Different from them, \sysname targets a new scenario, LLM inference serving.

\parabf{Inference serving.}
Many traditional model serving systems~\cite{olston2017tensorflow, nvidiatriton, crankshaw2017clipper, gujarati2020serving, zhangshepherd} only focus on serving relatively small models in a cluster without awareness of characteristics of LLMs. 
Recently, several serving systems are
proposed to optimize Transformer-based LLMs~\cite{fang2021turbotransformers, li2023mpcformer, li2023alpaserve, yu2022orca, vtc}. Orca~\cite{yu2022orca} and vLLM~\cite{vllm} considers the autoregressive generation pattern of LLMs. However,
due to their FCFS policy, they suffer from severe head-of-line blocking problem.
\revisioneurosys{VTC~\cite{vtc} focus on the fairness of LLM serving but does not consider the preemption scenario.
Splitwise~\cite{patel2024splitwise} and DistServe~\cite{zhong2024distserve} disaggregates the prefill and decoding phase to eliminate the interference between them and thus optimize execution latency. LoongServe~\cite{wu2024loongserve} uses elastic sequence parallelism to dynamically set degree of parallelism for different requests at different phases. These systems are orthogonal to \sysname.}

\parabf{Memory optimization for LLMs.} Due to high memory usage for LLMs, many techniques have been proposed to reduce memory overhead. Some work~\cite{gradientcompression, wang2023finetuning} targets training, which is orthogonal to the serving scenario. Quantization~\cite{li2020train, xiao2022smoothquant, frantar2022gptq,
dettmers2022llm} compresses the model weights into lower precision after training to reduce the memory footprint during inference. SparTA~\cite{sparta} exploits model sparsity to accelerate computation. However, these
approaches sacrifice the model accuracy.
vLLM~\cite{vllm} proposes PagedAttention to reduce the GPU memory fragmentation. This is orthogonal to this paper and \sysname implements PagedAttention as well.

\section{Conclusion}
\label{sec:conclusion}

We present \sysname, a distributed inference serving system for
LLMs. We exploit the autoregressive pattern
of LLM inference to enable iteration-level preemption and design a novel
skip-join MLFQ scheduler to address head-of-line blocking problem. We propose a
proactive key-value cache management mechanism to handle the memory overhead of
the key-value cache and hide the data transmission latency with computing. Based
on these, we build a prototype of \sysname. Experiments show that
\sysname improves the throughput by up to 31.4$\times$ and
17.9$\times$ under the same average and tail latency SLOs respectively, compared to vLLM.
\label{lastpage}

\bibliographystyle{ieeetr}
\bibliography{paper}

\begin{thebibliography}{10}

\bibitem{chatgpt}
``{Introducing ChatGPT}.'' \url{https://openai.com/blog/chatgpt}, 2022.

\bibitem{chatgptnews}
``{ChatGPT sets record for fastest-growing user base}.'' \url{https://www.reuters.com/technology/chatgpt-sets-record-fastest-growing-user-base-analyst-note-2023-02-01/}, 2023.

\bibitem{newbing}
``Reinventing search with a new ai-powered bing and edge, your copilot for the web.'' \url{https://news.microsoft.com/the-new-Bing/}, 2023.

\bibitem{gemini}
Google, ``{Our next-generation model: Gemini 1.5}.'' \url{https://blog.google/technology/ai/google-gemini-next-generation-model-february-2024/}, 2024.

\bibitem{claude}
Anthropic, ``{Introducing the next generation of Claude}.'' \url{https://www.anthropic.com/news/claude-3-family}, 2024.

\bibitem{qwen}
``{Introducing Qwen}.'' \url{https://qwenlm.github.io/blog/qwen/}, 2023.

\bibitem{he2016deep}
K.~He, X.~Zhang, S.~Ren, and J.~Sun, ``Deep residual learning for image recognition,'' in {\em IEEE Conference on Computer Vision and Pattern Recognition}, 2016.

\bibitem{gujarati2020serving}
A.~Gujarati, R.~Karimi, S.~Alzayat, W.~Hao, A.~Kaufmann, Y.~Vigfusson, and J.~Mace, ``Serving {DNNs} like clockwork: Performance predictability from the bottom up,'' in {\em USENIX OSDI}, 2020.

\bibitem{zhangshepherd}
H.~Zhang, Y.~Tang, A.~Khandelwal, and I.~Stoica, ``Shepherd: Serving dnns in the wild,'' in {\em USENIX NSDI}, 2023.

\bibitem{yu2022orca}
G.-I. Yu, J.~S. Jeong, G.-W. Kim, S.~Kim, and B.-G. Chun, ``Orca: A distributed serving system for {Transformer-Based} generative models,'' in {\em USENIX OSDI}, 2022.

\bibitem{vllm}
W.~Kwon, Z.~Li, S.~Zhuang, Y.~Sheng, L.~Zheng, C.~H. Yu, J.~E. Gonzalez, H.~Zhang, and I.~Stoica, ``Efficient memory management for large language model serving with pagedattention,'' in {\em ACM SOSP}, 2023.

\bibitem{shinjuku}
K.~Kaffes, T.~Chong, J.~T. Humphries, A.~Belay, D.~Mazi{\`e}res, and C.~Kozyrakis, ``Shinjuku: Preemptive scheduling for $\mu$second-scale tail latency,'' in {\em USENIX NSDI}, 2019.

\bibitem{sharegpt}
``Sharegpt teams.'' \url{https://sharegpt.com/}, 2023.

\bibitem{alpaca}
R.~Taori, I.~Gulrajani, T.~Zhang, Y.~Dubois, X.~Li, C.~Guestrin, P.~Liang, and T.~B. Hashimoto, ``Stanford alpaca: An instruction-following llama model.'' \url{https://github.com/tatsu-lab/stanford_alpaca}, 2023.

\bibitem{bai2015information}
W.~Bai, L.~Chen, K.~Chen, D.~Han, C.~Tian, and H.~Wang, ``Information-agnostic flow scheduling for commodity data centers,'' in {\em USENIX OSDI}, 2015.

\bibitem{shoeybi2020megatronlm}
M.~Shoeybi, M.~Patwary, R.~Puri, P.~LeGresley, J.~Casper, and B.~Catanzaro, ``Megatron-lm: Training multi-billion parameter language models using model parallelism,'' {\em arXiv}, 2020.

\bibitem{huang2019gpipe}
Y.~Huang, Y.~Cheng, A.~Bapna, O.~Firat, M.~X. Chen, D.~Chen, H.~Lee, J.~Ngiam, Q.~V. Le, Y.~Wu, and Z.~Chen, ``Gpipe: Efficient training of giant neural networks using pipeline parallelism,'' {\em Neural Information Processing Systems}, 2019.

\bibitem{zhang2022opt}
S.~Zhang, S.~Roller, N.~Goyal, M.~Artetxe, M.~Chen, S.~Chen, C.~Dewan, M.~Diab, X.~Li, X.~V. Lin, T.~Mihaylov, M.~Ott, S.~Shleifer, K.~Shuster, D.~Simig, P.~S. Koura, A.~Sridhar, T.~Wang, and L.~Zettlemoyer, ``Opt: Open pre-trained transformer language models,'' {\em arXiv}, 2022.

\bibitem{brown2020language}
T.~B. Brown, B.~Mann, N.~Ryder, M.~Subbiah, J.~Kaplan, P.~Dhariwal, A.~Neelakantan, P.~Shyam, G.~Sastry, A.~Askell, S.~Agarwal, A.~Herbert-Voss, G.~Krueger, T.~Henighan, R.~Child, A.~Ramesh, D.~M. Ziegler, J.~Wu, C.~Winter, C.~Hesse, M.~Chen, E.~Sigler, M.~Litwin, S.~Gray, B.~Chess, J.~Clark, C.~Berner, S.~McCandlish, A.~Radford, I.~Sutskever, and D.~Amodei, ``Language models are few-shot learners,'' {\em arXiv}, 2020.

\bibitem{touvron2023llama}
H.~Touvron, T.~Lavril, G.~Izacard, X.~Martinet, M.-A. Lachaux, T.~Lacroix, B.~Rozière, N.~Goyal, E.~Hambro, F.~Azhar, A.~Rodriguez, A.~Joulin, E.~Grave, and G.~Lample, ``Llama: Open and efficient foundation language models,'' {\em arXiv}, 2023.

\bibitem{vaswani2017attention}
A.~Vaswani, N.~Shazeer, N.~Parmar, J.~Uszkoreit, L.~Jones, A.~N. Gomez, {\L}.~Kaiser, and I.~Polosukhin, ``Attention is all you need,'' {\em Neural Information Processing Systems}, 2017.

\bibitem{olston2017tensorflow}
C.~Olston, N.~Fiedel, K.~Gorovoy, J.~Harmsen, L.~Lao, F.~Li, V.~Rajashekhar, S.~Ramesh, and J.~Soyke, ``Tensorflow-serving: Flexible, high-performance ml serving,'' {\em arXiv}, 2017.

\bibitem{nvidiatriton}
N.~Corporation, ``Triton inference server: An optimized cloud and edge inferencing solution.,'' 2019.

\bibitem{ott2019fairseq}
M.~Ott, S.~Edunov, A.~Baevski, A.~Fan, S.~Gross, N.~Ng, D.~Grangier, and M.~Auli, ``fairseq: A fast, extensible toolkit for sequence modeling,'' {\em arXiv}, 2019.

\bibitem{wolf2020huggingfaces}
T.~Wolf, L.~Debut, V.~Sanh, J.~Chaumond, C.~Delangue, A.~Moi, P.~Cistac, T.~Rault, R.~Louf, M.~Funtowicz, J.~Davison, S.~Shleifer, P.~von Platen, C.~Ma, Y.~Jernite, J.~Plu, C.~Xu, T.~L. Scao, S.~Gugger, M.~Drame, Q.~Lhoest, and A.~M. Rush, ``Huggingface's transformers: State-of-the-art natural language processing,'' {\em arXiv}, 2020.

\bibitem{fastertransformer}
N.~Corporation, ``Fastertransformer,'' 2019.

\bibitem{schrage1968proof}
L.~Schrage, ``A proof of the optimality of the shortest remaining processing time discipline,'' {\em Operations Research}, 1968.

\bibitem{shazeer2019fast}
N.~Shazeer, ``Fast transformer decoding: One write-head is all you need,'' {\em arXiv}, 2019.

\bibitem{gqa}
J.~Ainslie, J.~Lee-Thorp, M.~de~Jong, Y.~Zemlyanskiy, F.~Lebrón, and S.~Sanghai, ``Gqa: Training generalized multi-query transformer models from multi-head checkpoints,'' {\em arXiv}, 2023.

\bibitem{longchat2023}
D.~Li*, R.~Shao*, A.~Xie, Y.~Sheng, L.~Zheng, J.~E. Gonzalez, I.~Stoica, X.~Ma, and H.~Zhang, ``How long can open-source llms truly promise on context length?,'' 2023.

\bibitem{flowpreempt}
C.-Y. Hong, M.~Caesar, and P.~B. Godfrey, ``Finishing flows quickly with preemptive scheduling,'' in {\em ACM SIGCOMM}, 2012.

\bibitem{alizadeh2013pfabric}
M.~Alizadeh, S.~Yang, M.~Sharif, S.~Katti, N.~McKeown, B.~Prabhakar, and S.~Shenker, ``pfabric: Minimal near-optimal datacenter transport,'' {\em SIGCOMM CCR}, 2013.

\bibitem{chowdhury2015efficient}
M.~Chowdhury and I.~Stoica, ``Efficient coflow scheduling without prior knowledge,'' {\em SIGCOMM CCR}, 2015.

\bibitem{gu2019tiresias}
J.~Gu, M.~Chowdhury, K.~G. Shin, Y.~Zhu, M.~Jeon, J.~Qian, H.~H. Liu, and C.~Guo, ``Tiresias: A gpu cluster manager for distributed deep learning.,'' in {\em USENIX NSDI}, 2019.

\bibitem{li2023alpaserve}
Z.~Li, L.~Zheng, Y.~Zhong, V.~Liu, Y.~Sheng, X.~Jin, Y.~Huang, Z.~Chen, H.~Zhang, J.~E. Gonzalez, {\em et~al.}, ``{AlpaServe}: Statistical multiplexing with model parallelism for deep learning serving,'' in {\em USENIX OSDI}, 2023.

\bibitem{kaplan2020scaling}
J.~Kaplan, S.~McCandlish, T.~Henighan, T.~B. Brown, B.~Chess, R.~Child, S.~Gray, A.~Radford, J.~Wu, and D.~Amodei, ``Scaling laws for neural language models,'' {\em arXiv}, 2020.

\bibitem{narayanan2021efficient}
D.~Narayanan, M.~Shoeybi, J.~Casper, P.~LeGresley, M.~Patwary, V.~A. Korthikanti, D.~Vainbrand, P.~Kashinkunti, J.~Bernauer, B.~Catanzaro, A.~Phanishayee, and M.~Zaharia, ``Efficient large-scale language model training on gpu clusters using megatron-lm,'' {\em arXiv}, 2021.

\bibitem{pipedream}
D.~Narayanan, A.~Harlap, A.~Phanishayee, V.~Seshadri, N.~R. Devanur, G.~R. Ganger, P.~B. Gibbons, and M.~Zaharia, ``Pipedream: Generalized pipeline parallelism for dnn training,'' in {\em ACM SOSP}, 2019.

\bibitem{ray}
P.~Moritz, R.~Nishihara, S.~Wang, A.~Tumanov, R.~Liaw, E.~Liang, M.~Elibol, Z.~Yang, W.~Paul, M.~I. Jordan, and I.~Stoica, ``Ray: A distributed framework for emerging {AI} applications,'' in {\em USENIX OSDI}, 2018.

\bibitem{zygos}
G.~Prekas, M.~Kogias, and E.~Bugnion, ``Zygos: Achieving low tail latency for microsecond-scale networked tasks,'' in {\em ACM SOSP}, 2017.

\bibitem{zhong2024distserve}
Y.~Zhong, S.~Liu, J.~Chen, J.~Hu, Y.~Zhu, X.~Liu, X.~Jin, and H.~Zhang, ``Distserve: Disaggregating prefill and decoding for goodput-optimized large language model serving,'' in {\em USENIX OSDI}, 2024.

\bibitem{wu2024loongserve}
B.~Wu, S.~Liu, Y.~Zhong, P.~Sun, X.~Liu, and X.~Jin, ``Loongserve: Efficiently serving long-context large language models with elastic sequence parallelism,'' {\em arXiv}, 2024.

\bibitem{chowdhury2014efficient}
M.~Chowdhury, Y.~Zhong, and I.~Stoica, ``Efficient coflow scheduling with varys,'' in {\em ACM SIGCOMM}, 2014.

\bibitem{kaffes2019shinjuku}
K.~Kaffes, T.~Chong, J.~T. Humphries, A.~Belay, D.~Mazi{\`e}res, and C.~Kozyrakis, ``Shinjuku: Preemptive scheduling for $\mu$second-scale tail latency,'' in {\em USENIX NSDI}, 2019.

\bibitem{ousterhout2019shenango}
A.~Ousterhout, J.~Fried, J.~Behrens, A.~Belay, and H.~Balakrishnan, ``Shenango: Achieving high cpu efficiency for latency-sensitive datacenter workloads.,'' in {\em USENIX NSDI}, 2019.

\bibitem{fried2020caladan}
J.~Fried, Z.~Ruan, A.~Ousterhout, and A.~Belay, ``Caladan: Mitigating interference at microsecond timescales,'' in {\em USENIX OSDI}, 2020.

\bibitem{bai2020pipeswitch}
Z.~Bai, Z.~Zhang, Y.~Zhu, and X.~Jin, ``Pipeswitch: Fast pipelined context switching for deep learning applications,'' in {\em USENIX OSDI}, 2020.

\bibitem{preemptionInference}
M.~Han, H.~Zhang, R.~Chen, and H.~Chen, ``Microsecond-scale preemption for concurrent {GPU-accelerated} {DNN} inferences,'' in {\em USENIX OSDI}, 2022.

\bibitem{crankshaw2017clipper}
D.~Crankshaw, X.~Wang, G.~Zhou, M.~J. Franklin, J.~E. Gonzalez, and I.~Stoica, ``Clipper: A low-latency online prediction serving system.,'' in {\em USENIX NSDI}, 2017.

\bibitem{fang2021turbotransformers}
J.~Fang, Y.~Yu, C.~Zhao, and J.~Zhou, ``Turbotransformers: an efficient gpu serving system for transformer models,'' in {\em ACM PPoPP}, 2021.

\bibitem{li2023mpcformer}
D.~Li, R.~Shao, H.~Wang, H.~Guo, E.~P. Xing, and H.~Zhang, ``Mpcformer: fast, performant and private transformer inference with mpc,'' {\em arXiv}, 2023.

\bibitem{vtc}
Y.~Sheng, S.~Cao, D.~Li, B.~Zhu, Z.~Li, D.~Zhuo, J.~E. Gonzalez, and I.~Stoica, ``Fairness in serving large language models,'' in {\em USENIX OSDI}, 2024.

\bibitem{patel2024splitwise}
P.~Patel, E.~Choukse, C.~Zhang, A.~Shah, Íñigo Goiri, S.~Maleki, and R.~Bianchini, ``Splitwise: Efficient generative llm inference using phase splitting,'' in {\em ACM/IEEE ISCA}, 2024.

\bibitem{gradientcompression}
Y.~Bai, C.~Li, Q.~Zhou, J.~Yi, P.~Gong, F.~Yan, R.~Chen, and Y.~Xu, ``Gradient compression supercharged high-performance data parallel dnn training,'' in {\em ACM SOSP}, 2021.

\bibitem{wang2023finetuning}
J.~Wang, B.~Yuan, L.~Rimanic, Y.~He, T.~Dao, B.~Chen, C.~Re, and C.~Zhang, ``Fine-tuning language models over slow networks using activation quantization with guarantees,'' {\em Neural Information Processing Systems}, 2022.

\bibitem{li2020train}
Z.~Li, E.~Wallace, S.~Shen, K.~Lin, K.~Keutzer, D.~Klein, and J.~Gonzalez, ``Train big, then compress: Rethinking model size for efficient training and inference of transformers,'' in {\em International Conference on Machine Learning (ICML)}, 2020.

\bibitem{xiao2022smoothquant}
G.~Xiao, J.~Lin, M.~Seznec, J.~Demouth, and S.~Han, ``Smoothquant: Accurate and efficient post-training quantization for large language models,'' {\em International Conference on Machine Learning}, 2022.

\bibitem{frantar2022gptq}
E.~Frantar, S.~Ashkboos, T.~Hoefler, and D.~Alistarh, ``Gptq: Accurate post-training quantization for generative pre-trained transformers,'' {\em arXiv}, 2022.

\bibitem{dettmers2022llm}
T.~Dettmers, M.~Lewis, Y.~Belkada, and L.~Zettlemoyer, ``Llm. int8 (): 8-bit matrix multiplication for transformers at scale,'' {\em arXiv}, 2022.

\bibitem{sparta}
N.~Zheng, B.~Lin, Q.~Zhang, L.~Ma, Y.~Yang, F.~Yang, Y.~Wang, M.~Yang, and L.~Zhou, ``{SparTA}: {Deep-Learning} model sparsity via {Tensor-with-Sparsity-Attribute},'' in {\em USENIX OSDI}, 2022.

\end{thebibliography}

\clearpage

\end{document}